\documentclass[acmsmall]{acmart}
\usepackage{multirow}
\usepackage{booktabs}
\usepackage{pifont,makecell}
\AtBeginDocument{%
  }

\usepackage{hyperref}
\usepackage{bm}
\setcopyright{acmlicensed}
\copyrightyear{2018}
\acmYear{2018}
\acmDOI{XXXXXXX.XXXXXXX}

\acmJournal{JACM}
\acmVolume{37}
\acmNumber{4}
\acmArticle{111}
\acmMonth{8}

\begin{document}

\title{Benchmarking Multi-dimensional AIGC Video Quality Assessment: A Dataset and Unified Model}




\author{Zhichao Zhang}
\email{liquortect@sjtu.edu.cn}
\author{Wei Sun}
\email{sunguwei@sjtu.edu.cn}
\author{Xinyue Li}
\email{xinyueli@sjtu.edu.cn}
\author{Jun Jia}
\email{jiajun0302@sjtu.edu.cn}
\author{Xiongkuo Min}
\email{minxiongkuo@sjtu.edu.cn}
\author{Zicheng Zhang}
\email{zzc1998@sjtu.edu.cn}
\author{Chunyi Li}
\email{lcysyzxdxc@sjtu.edu.cn}
\author{Zijian Chen}
\email{zijian.chen@sjtu.edu.cn}
\author{Puyi Wang}
\email{wangpuyi@sjtu.edu.cn}
\affiliation{%
  \institution{Shanghai Jiao Tong University}
  \city{Shanghai}
  \country{China}}

\author{Fengyu Sun}
\email{sunfengyu@hisilicon.com}
\author{Shangling Jui}
\email{jui.shangling@huawei.com}
\affiliation{%
  \institution{Huawei Technologies}
  \city{Shanghai}
  \country{China}}

\author{Guangtao Zhai}
\email{zhaiguangtao@sjtu.edu.cn}
\affiliation{%
  \institution{Shanghai Jiao Tong University}
  \city{Shanghai}
  \country{China}}

\renewcommand{\shortauthors}{Trovato et al.}

\begin{abstract}
In recent years, \textbf{artificial intelligence (AI)}-driven video generation has gained significant attention due to great advancements in visual and language generative techniques. Consequently, there is a growing need for accurate \textbf{video quality assessment (VQA)} metrics to evaluate the perceptual quality of \textbf{AI-generated content (AIGC)} videos and optimize video generation models. However, assessing the quality of AIGC videos remains a significant challenge because these videos often exhibit highly complex distortions, such as unnatural actions and irrational objects. 
To address this challenge, we systematically investigate the AIGC-VQA problem in this paper, considering both subjective and objective quality assessment perspectives. For the subjective perspective, we construct the \textbf{\underline{L}arge-scale \underline{G}enerated \underline{V}ideo \underline{Q}uality assessment} \textbf{(LGVQ)} dataset, consisting of $2,808$ AIGC videos generated by six video generation models using $468$ carefully curated text prompts. Unlike previous subjective VQA experiments, we evaluate the perceptual quality of AIGC videos from three critical dimensions: \textbf{spatial quality, temporal quality, and text-video alignment}, which hold utmost importance for current video generation techniques. 
For the objective perspective, we establish a benchmark for evaluating existing quality assessment metrics on the LGVQ dataset. Our findings show that current metrics perform poorly on this dataset, highlighting a gap in effective evaluation tools. To bridge this gap, we propose the \textbf{\underline{U}nify \underline{G}enerated \underline{V}ideo \underline{Q}uality assessment} \textbf{(UGVQ)} model, designed to accurately evaluate the multi-dimensional quality of AIGC videos. The UGVQ model integrates the visual and motion features of videos with the textual features of their corresponding prompts, forming a unified quality-aware feature representation tailored to AIGC videos. Experimental results demonstrate that UGVQ achieves state-of-the-art performance on the LGVQ dataset across all three quality dimensions, validating its effectiveness as an accurate quality metric for AIGC videos.
We hope that our benchmark can promote the development of AIGC-VQA studies. Both the LGVQ dataset and the UGVQ model are publicly available on \href{https://github.com/zczhang-sjtu/UGVQ.git}{https://github.com/zczhang-sjtu/UGVQ.git}.
\end{abstract}




\received{20 February 2007}
\received[revised]{12 March 2009}
\received[accepted]{5 June 2009}

\maketitle

\section{Introduction}
\label{sec_introduction}

\begin{figure*}
    \centering
    \includegraphics[width=0.99\linewidth]{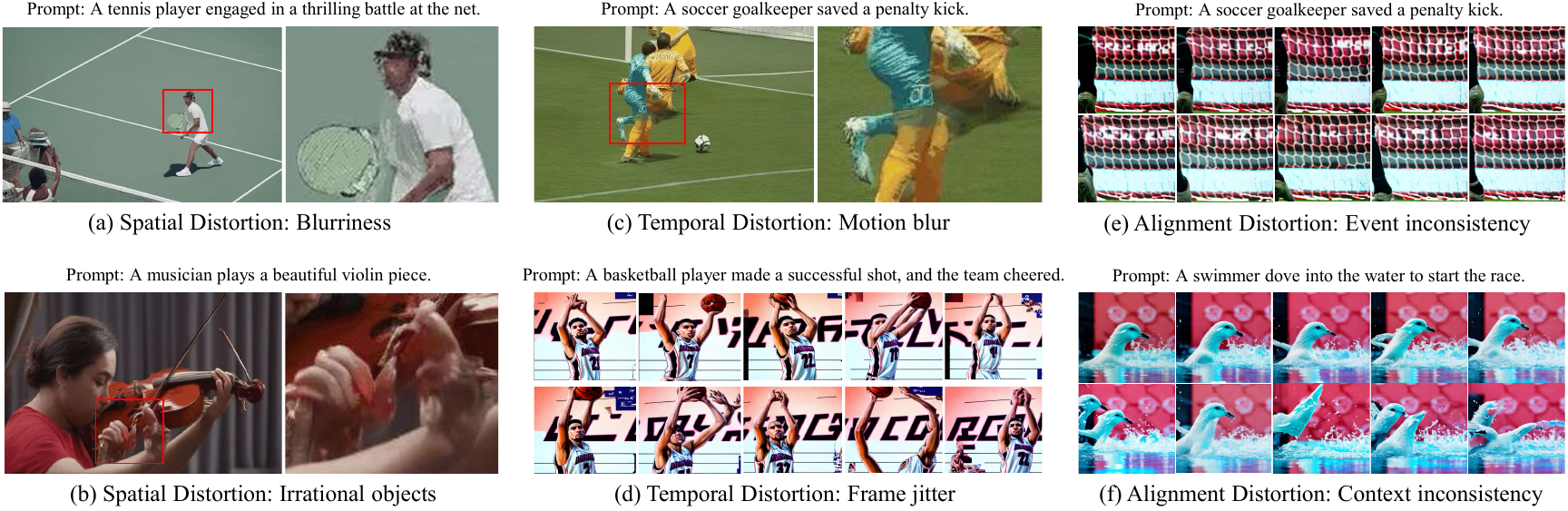}
    \caption{
    Typical distortion types of AIGC videos. Spatial distortions mainly include (a) blurriness and (b) irrational objects; temporal distortions include (c) motion blur and (d) frame jitter; alignment distortions include (e) event inconsistency and (f) context inconsistency.
    }
    \label{fig_distortion}
\end{figure*}

\textbf{Artificial intelligence (AI)}-generated video techniques, represented by ~\textbf{Text-to-Video (T2V)}~\cite{VideoCrafter, Text2Video, Tune-A-Video}, have garnered significant attention in recent years. 
Due to the highly simplistic generation process (\textit{e.g.} videos are generated entirely from textual descriptions), ~\textbf{AI-generated content} (AIGC) videos~\footnote{In the following paper, we also use T2V videos to refer to AIGC videos.} have found extensive applications in industries including film~\cite{aigc_film}, gaming~\cite{aigc_game}, advertising~\cite{aigc_ad}, and more. Despite significant progress in the development of AIGC video techniques, various quality challenges remain. As illustrated in Fig.~\ref{fig_distortion}, AIGC videos often exhibit notable spatial and temporal distortions, such as blurred objects, inconsistent backgrounds, and poor action continuity. Furthermore, misalignments between AIGC videos and their corresponding textual descriptions can hinder their practical applicability. Therefore, how to effectively evaluate the perceptual quality of AIGC videos is crucial for measuring the progress of video generation techniques, selecting the best AIGC videos from a set of candidates generated by T2V models, and optimizing the video generation techniques~\cite{black2023training}.




In previous video generation studies~\cite{VideoCrafter, Text2Video, Hotshot, Tune-A-Video}, only a few metrics have been utilized to evaluate the effectiveness of video generation models, such as ~\textbf{Inception Score (IS)}~\cite{IS}, ~\textbf{Fréchet Inception Distance} (FID)~\cite{FID}, ~\textbf{Fréchet Video Distance (FVD)}~\cite{FVD}, ~\textbf{Kernel Video Distance (KVD)}~\cite{KVD}, CLIPScore~\cite{CLIPScore}. IS assesses the quality and diversity of generated content by analyzing the confidence and diversity of label probabilities from a pre-trained classifier, but it is limited by its reliance on specific models, and inability to capture temporal or contextual coherence in videos. FID, FVD, and KVD compare the distribution of Inception~\cite{inception} features of generated frames with that of a set of real images/videos, thus failing to capture distortion-level and semantic-level quality characteristics. Furthermore, motion generation poses a great challenge for current video generation techniques, yet FID~\cite{FID} and FVD~\cite{FVD} are unable to quantify the impact of temporal-level distortions on visual quality. CLIP-based methods such as CLIPScore~\cite{CLIPScore}, and BLIP~\cite{BLIP} are frequently employed to assess the alignment between the generated video and its prompt text. However, CLIP-based methods can only assess frame-level alignment between the video frames and the text prompt, and they cannot also evaluate the alignment of videos containing diverse motions. Given these limitations, it is doubt to rely on existing metrics to measure progress in video generation techniques. Hence, it is necessary to investigate the extent to which current metrics can effectively evaluate the generation quality of AIGC videos.

Towards these goals, we systematically investigate the AIGC-VQA problem from both subjective and objective quality assessment perspectives. First,
given the absence of subjective AIGC-VQA datasets to serve as a benchmark, we construct a \textbf{\underline{L}arge-scale \underline{G}enerated \underline{V}ideo \underline{Q}uality assessment} \textbf{(LGVQ)} dataset to subjectively evaluate the three quality dimensions of AIGC videos---\textbf{spatial quality, temporal quality, and text-video alignment}---addressing the primary concerns of current video generation techniques. To make the generated video content encompasses a broad range of real-world scenarios, we structurally divide the text prompts into three components: \textbf{foreground content, background content, and motion state}. For each component, we include typical elements that frequently occur in daily life. Specifically, the foreground content includes four categories: \textbf{people}, \textbf{animals}, \textbf{plants}, and \textbf{man-made objects}, and the background content is categorized into \textbf{indoor scenes}, \textbf{outdoor natural scenes}, and \textbf{outdoor man-made scenes}. Motion state focuses on three types: \textbf{static}, \textbf{dynamic}, and \textbf{local movement} (\textit{e.g., watching belongs to static, running belongs to dynamic, and taking belongs to local movement}). By combining different words or phrases from the aforementioned types, we obtain $468$ text prompts. We use six mainstream T2V algorithms to generate $2,808$ AIGC videos using these text prompts. We then invite $60$ subjects to provide their perceptual ratings of the spatial quality, temporal quality, and text-video alignment for each video. We calculate the ~\textbf{mean opinion scores (MOSs)} as their quality labels.


Subsequently, we benchmark existing quality metrics on the LGVQ dataset to analyze their ability to assess the quality of AIGC videos. Since LGVQ provides spatial quality, temporal quality, and text-video alignment labels for each AIGC video, we include three categories of quality assessment metrics in the AIGC-VQA benchmark: $14$ ~\textbf{image quality assessment (IQA)} methods for spatial quality evaluation, $16$ ~\textbf{video quality assessment (VQA)} methods for temporal quality evaluation, and $9$ \textbf{CLIP-based and visual question-answering-based methods} for text-video alignment evaluation. The benchmark results reveal that none of the current quality metrics sufficiently assess one or more quality aspects of AIGC videos. This suggests that current metrics are not suitable for evaluating the quality of AIGC videos and the progress of the video generation techniques, thus highlighting the urgent need for effective AIGC VQA metrics.


To bridge this gap, we propose the \textbf{\underline{U}nify \underline{G}enerated \underline{V}ideo \underline{Q}uality assessment} \textbf{(UGVQ)} model to comprehensively evaluate the quality of AIGC videos. UGVQ systematically extract the spatial, temporal, and textual features, integrating them into a unified quality-aware feature representation for AIGC videos. For spatial features, we pre-trained a ~\textbf{Vision Transformer (ViT)} model~\cite{vit} on a large-scale AIGC IQA dataset, Pick-a-pic~\cite{PickScore}, to learn quality-aware features for detecting AIGC-related artifacts. The pre-trained ViT extracts spatial features from key frames, while a Transformer encoder aggregates these frame-level features into a video-level spatial quality representation. For temporal features, we utilize SlowFast~\cite{SlowFast}, a powerful action recognition model, to capture motion-related quality representation. 
To evaluate text-video alignment, we leverage the textual encoder of CLIP~\cite{CLIP} to extract semantic content representation from the text prompts associated with the videos.
We propose a multi-modality feature fusion module to integrate and refine the spatial, temporal, and textual features into the unified quality-aware representation. Finally, we perform a ~\textbf{multilayer perceptron (MLP)} network to regress these features into multi-dimensional quality scores.
Experimental results demonstrate that UGVQ achieves superior performance in evaluating all three quality dimensions of AIGC videos compared to existing quality metrics re-trained on the LGVQ dataset, indicating that UGVQ is an effective and comprehensive VQA metric for AIGC videos.


In summary, the contributions are summarized as:
\begin{itemize}
\item We construct the \textbf{Large-scale Generated Video Quality assessment} \textbf{(LGVQ)} dataset and conduct a subjective VQA experiment to derive spatial quality, temporal quality, and text-video alignment labels for each AIGC video. The diversity of LGVQ and multi-dimensional quality labels make it well-suited for validating and developing VQA models for AIGC videos.
\item We establish a comprehensive benchmark for AIGC-VQA, which verifies the performance of existing IQA, VQA, and alignment evaluation metrics in evaluating the quality of AIGC videos.
\item We develop the \textbf{Unify Generated Video Quality assessment} \textbf{(UGVQ)} model to effectively evaluate the AIGC videos across spatial quality, temporal quality, and text-video alignment dimensions. The experimental results demonstrate that UGVQ is superior to other quality metrics. 
\end{itemize}

\section{Related Works}
\label{sec_related_works}
\subsection{Video Generation Techniques}
Text-to-video generation can be categorized into three frameworks: \textbf{Variational Autoencoders (VAE)} or \textbf{Generative Adversarial Networks (GAN)-based}~\cite{li2018video,deng2019irc}, \textbf{Autoregressive-based}~\cite{wu2022nuwa,liang2022nuwa,hong2022cogvideo,villegas2022phenaki}, and \textbf{Diffusion-based}~\cite{ho2022video,he2022latent,wu2023tune,chen2023videocrafter1,yin2023nuwa}.

Early research on T2V models predominantly employed \textbf{VAE/GAN} for video generation. For instance, Li \textit{et al.}~\cite{li2018video} trained a conditional video generative model combining VAE and GAN to extract static and dynamic information from text. Deng \textit{et al.}~\cite{deng2019irc} proposed the introspective recurrent convolutional GAN, integrating 2D transconvolutional layers with LSTM cells to ensure temporal coherence and semantically align generated videos with input text.

Subsequently, \textbf{autoregressive models} have also been explored for T2V generation. For instance, N\"UWA~\cite{wu2022nuwa} leverages a 3D transformer encoder-decoder with a nearby attention mechanism for high-quality video synthesis. N\"UWA-Infinity~\cite{liang2022nuwa} presents a ``render-and-optimize” strategy for infinite visual generation. CogVideo~\cite{hong2022cogvideo} utilizes pre-trained weights from the text-to-image model and employs a multi-frame-rate hierarchical training strategy to enhance text-video alignment. Phenaki~\cite{villegas2022phenaki} uses a variable-length video generation method with a C-ViViT encoder-decoder structure to compress video into discrete tokens.

Recently, \textbf{diffusion models}~\cite{ho2020denoising} have significantly advanced T2V generation and have emerged as the dominant architecture for T2V generation~\cite{betker2023improving,zhang2023adding,zhang2023text}. The Video Diffusion Model~\cite{ho2022video} applies the diffusion model to video generation using a 3D U-Net architecture combined with temporal attention. To reduce computational complexity, LVDM~\cite{he2022latent} introduces a hierarchical latent video diffusion model. Gen-1~\cite{esser2023structure} is a structure and content-guided video diffusion model, training on monocular depth estimates for control over structure and content. Tune-a-video~\cite{wu2023tune} employs a spatiotemporal attention mechanism to maintain frame consistency. Video Crafter1~\cite{chen2023videocrafter1} uses a video VAE and a video latent diffusion process for lower-dimensional latent representation and video generation. N\"UWA-XL~\cite{yin2023nuwa} uses two diffusion models to generate keyframes and refine adjacent frames.

Despite these advancements, challenges such as irregular human and object appearances, inconsistent motion,  and unrealistic background persist~\cite{sunsora,cho2024sora}. Assessing the quality of T2V videos is essential for measuring progress and further promoting the development of T2V models.

\subsection{Quality Metrics for AIGC Videos}

\textbf{Spatial quality metrics} aim to measure the frame-level visual quality of AIGC videos. IS~\cite{IS} and FID~\cite{FID} are the most frequently used metrics to evaluate spatial quality in the literature. However, many studies~\cite{toward,GAIA} have indicated that IS and FID exhibit poor correlation with human visual perception. On the other hand, image quality assessment is design to quantify the perceptual quality of images. Many popular IQA methods, such as SSIM~\cite{SSIM}, UNIQUE~\cite{UNIQUE}, StairIQA~\cite{StairIQA}, MUSIQ~\cite{MUSIQ}, LIQE~\cite{LIQE}, etc., have demonstrated remarkable capability in measuring the perceptual quality of natural images. Nevertheless, since these methods are trained on natural \textbf{scene content content (NSC)} image quality assessment datasets, there is uncertainty regarding their ability to evaluate the spatial quality of AIGC images/videos. Recently, AIGC-specific IQA methods, such as MA-AGIQA~\cite{MA_AGIQA} and IPCE~\cite{IPCE}, have been introduced. Through training on AIGC IQA datasets (e.g., AGIQA-3k~\cite{AGIQA_3K} and AIGCQA-20K~\cite{AIGIQA_20K}), these methods have shown promise in identifying AIGC-related artifacts. However, the spatial quality of video frames can also be influenced by temporal distortions~\cite{GAIA}. Consequently, it is essential to explore whether these NSC-based and AIGC-specific IQA methods can effectively evaluate the spatial quality of AIGC videos.

\textbf{Temporal quality metrics} are responsible for assessing the temporal coherence of AIGC videos.
Previous T2V studies utilize FVD~\cite{FVD} to gauge the disparity between features extracted by pre-trained ~\textbf{Inflated-3D Convnets (I3D)}~\cite{kinetics} from generated videos and realistic videos. 
Similar to FID, FVD also demonstrates a weak correlation with human visual perception. As related research, \textbf{user-generated content (UGC)} VQA models~\cite{min2024perceptual}, such as SimpleVQA~\cite{simpleVQA}, FastVQA~\cite{FAST_VQA}, DOVER~\cite{dover}, etc., have attempted to utilize action recognition network (\textit{e.g.} SlowFast~\cite{SlowFast}, Video Swin Transformer~\cite{Swin}) to represent the temporal quality feature. However, several studies~\cite{fang2023study,sun2024analysis} have shown that the current UGC VQA datasets pose little challenge to temporal quality analyzers in UGC VQA models, potentially rendering them ineffective in dealing with the complex temporal distortions in AIGC videos. Recently, several quality metrics specifically designed for AIGC videos have been proposed. For instance, EvalCrafter~\cite{EvalCrafter} validates 17 objective metrics, encompassing dimensions such as visual quality, text-video alignment, motion quality, and temporal consistency. Similarly, VBench~\cite{VBench} introduces 16 dimension metrics to systematically evaluate the video quality and video-condition consistency. However, these metrics are primarily designed for model-level evaluation of T2V systems and may lack precision for video-level quality assessment. \cite{T2VQA} develops an LMM-based metric, named T2VQA, which focuses on the overall quality of AIGC videos but cannot independently evaluate the quality of the temporal dimension.

\textbf{Text-video alignment metrics} evaluate the consistency between the generated videos and textual descriptions. CLIP-based methods, such as CLIP~\cite{CLIPScore}, BLIP~\cite{BLIP}, and viCLIP~\cite{viCLIP} are frequently used to evaluate the consistency between the AIGC videos and their text prompts. While these methods are trained on large-scale text-image datasets~\cite{PickScore,HPSv2} to maximize the similarity of positive pairs, recent studies~\cite{cogview2,toward} demonstrate that they have poor consistency with human visual perception. Hence, some studies have constructed human-rated text-image alignment datasets~\cite{ImageReward,PickScore,HPSv2}. Based on these datasets, they develop alignment assessment models, like ImageReward~\cite{ImageReward}, PickScore~\cite{PickScore}, HPSv1~\cite{HPSv1}, HPSv2~\cite{HPSv2}, etc., to evaluate the consistency between the text prompts and AIGC images. Beyond CLIP-based methods, recent researches~\cite{qalign,LMM_VQA} explore leveraging visual question-answering models for text-image alignment evaluation. The core idea involves creating question-answer pairs based on the text prompts and then applying visual question-answering models to the AIGC images to determine whether the models can provide correct answers. However, these approaches are primarily designed for text-image alignment, thus struggling to understand the motion concepts in the text and videos, making them insufficient to measure text-video alignment.

\section{Subjective Quality Assessment Study}
\label{sec_subjective}

\subsection{LGVQ Dataset}
We first construct LGVQ, a large-scale generated video dataset consisting of diverse AIGC videos, to serve as the benchmark for testing the perceptual quality of AIGC videos subjectively and objectively.


\begin{figure}
    \centering
    \begin{minipage}{0.48\textwidth}
        \centering
        \includegraphics[width=1\linewidth]{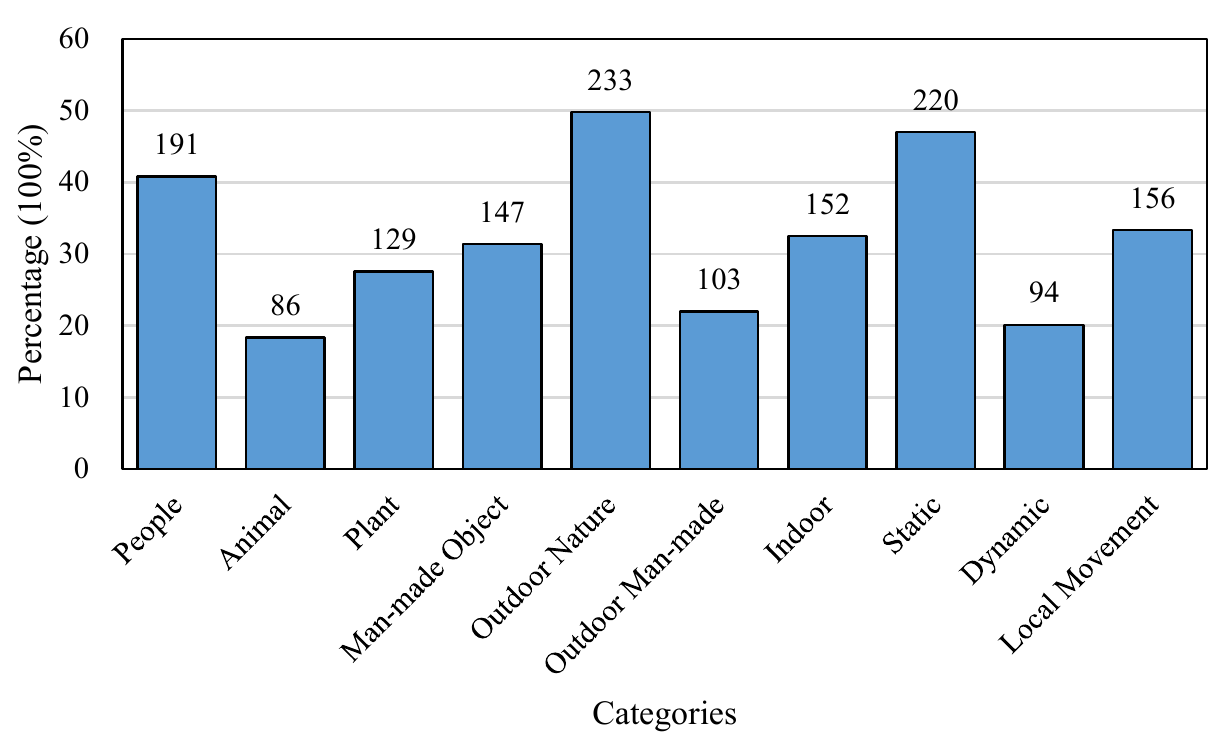} 
        \caption{The distribution of foreground content, background content, and motion state in LGVQ Dataset.}
        \label{fig_cls_distribution}
    \end{minipage}
    \hfill
    \begin{minipage}{0.48\textwidth}
        \centering
        \includegraphics[width=1\linewidth]{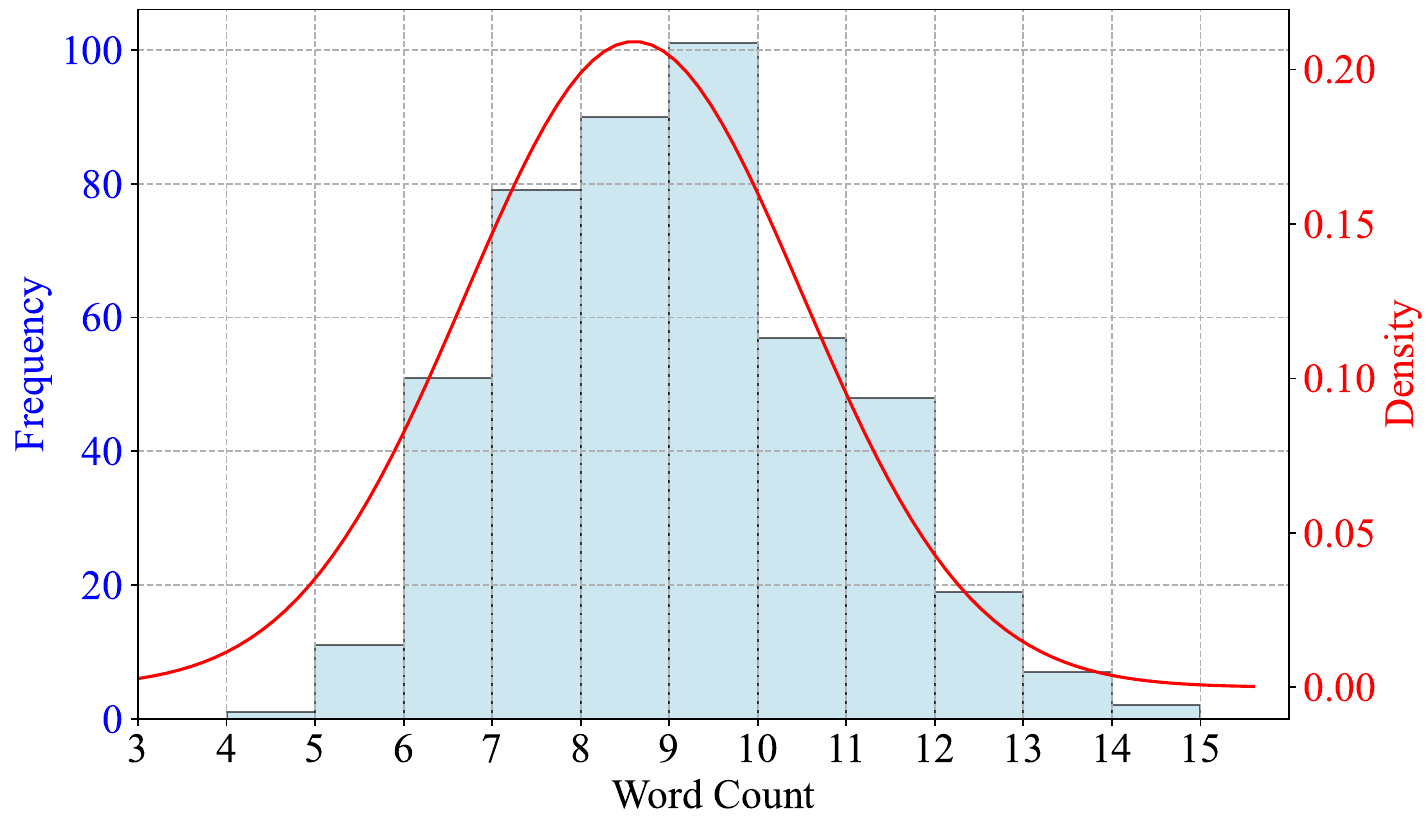} 
        \caption{The histogram and density plot of word length per prompt.}
        \label{fig_prompt_length}
    \end{minipage}
\end{figure}

\vspace{0.2cm}
\noindent\textbf{Prompts Selection.}
Text prompts are essential for describing the video content generated by T2V models. Consequently, the diversity of text prompts directly impacts the variety of video datasets. Existing AIGC datasets/benchmarks, such as VBench~\cite{VBench} and FETV~\cite{FETV}, provide relatively broad classifications. For example, VBench organizes text prompts into eight classes: \textbf{animal, architecture, food, human, lifestyle, plant, scenery,} and \textbf{vehicles}. Similarly, FETV outlines nine categories: \textbf{people, animals, vehicles, plants, artifacts, food, building, scenery,} and \textbf{illustrations}. While these categories offer a solid starting point, they lack specificity concerning motion and dynamic events, which is very important for video generation. 
To address this limitation, our dataset introduces a refined classification of motion types, dividing them into \textbf{static, dynamic, and local movement}. This detailed categorization allows for a more comprehensive representation of motion in real-world scenarios, enabling a systematic evaluation of T2V models' capabilities in generating videos with varying degrees of motion complexity.

Specifically, we decompose the text prompts into three components: \textbf{foreground content, background content, and motion state}. The \textbf{foreground content} represents the main subject of the event and includes four categories: \textbf{people, animals, plants, and man-made objects}. The \textbf{background content} refers to the environment or location where the event takes place, categorized into \textbf{indoor scenes, outdoor natural scenes, and outdoor man-made scenes}. The \textbf{motion state} defines the primary motion pattern of the event, classified into \textbf{static, dynamic, and local movement}. Each prompt needs to choose one or more words belonging to the components of \textbf{foreground content, background content}, and \textbf{motion state}. These words are then organized into complete sentences using GPT-4. Through this process, we generate a total of $468$ text prompts.
The distribution of foreground content, background content, and motion state in the LGVQ dataset are shown in Fig.~\ref{fig_cls_distribution}, showing that \textbf{people}, \textbf{outdoor natural scenes}, and \textbf{static} occupy the largest proportions in their respective components. The histogram density plot of word length per prompt in the LGVQ is shown in Fig.~\ref{fig_prompt_length}, which follows an approximately Gaussian distribution, with a maximum word length of $15$ and a minimum of $4$.


\begin{table}
\centering
  \caption{Video formats generated by the six T2V models in the LGVQ dataset.}
  \label{tab_frame}
  \resizebox{0.55\textwidth}{!}{
  \begin{tabular}{lcclll}
    \toprule
     Methods                           & License     & Duration (s)  & FPS  & Resolution         \\
    \midrule                                                                                       
     Gen-2~\cite{Gen2}                 & Commercial  & $4.0$          & $24$ & $1408 \times 768$  \\
     Tune-a-video~\cite{Tune-A-Video}  & Open-source & $2.4$          & $10$ & $512 \times 512$   \\
     Video Crafter~\cite{VideoCrafter} & Open-source & $2.0$          & $8$  & $256 \times 256$   \\
     Text2Video-Zero~\cite{Text2Video} & Open-source & $3.0$          & $12$ & $512 \times 512$   \\
     HotShot~\cite{Hotshot}            & Commercial  & $2.0$          & $4$  & $672 \times 384$   \\
     Video Fusion~\cite{VideoFusion}   & Open-source & $2.0$          & $8$  & $256 \times 256$   \\
    \bottomrule
  \end{tabular}
  }
\end{table}

\vspace{0.2cm}
\noindent\textbf{T2V Methods Selection.}
We select six popular text-to-video models, including Gen-2~\cite{Gen2}, Hotshot-XL~\cite{Hotshot}, Video Fusion~\cite{VideoFusion}, Video Crafter~\cite{VideoCrafter}, Text2Video-Zero~\cite{Text2Video}, and Tune-a-video~\cite{Tune-A-Video}, to generate videos for each prompt. The detailed formats of generated videos is shown in Table~\ref{tab_frame}. It can be observed that Gen-2 produces videos with the longest duration of $96$ frames, the highest frame rate of $24$ FPS, and the maximum resolution of $1408 \times 768$. In contrast, HotShot generates videos with the shortest duration of $8$ frames and the lowest frame rate of $4$ FPS. Additionally, both Video Crafter and Video Fusion produce videos with the lowest resolution of $256 \times 256$. 

\begin{figure*}
    \centering
    \includegraphics[width=0.9\linewidth]{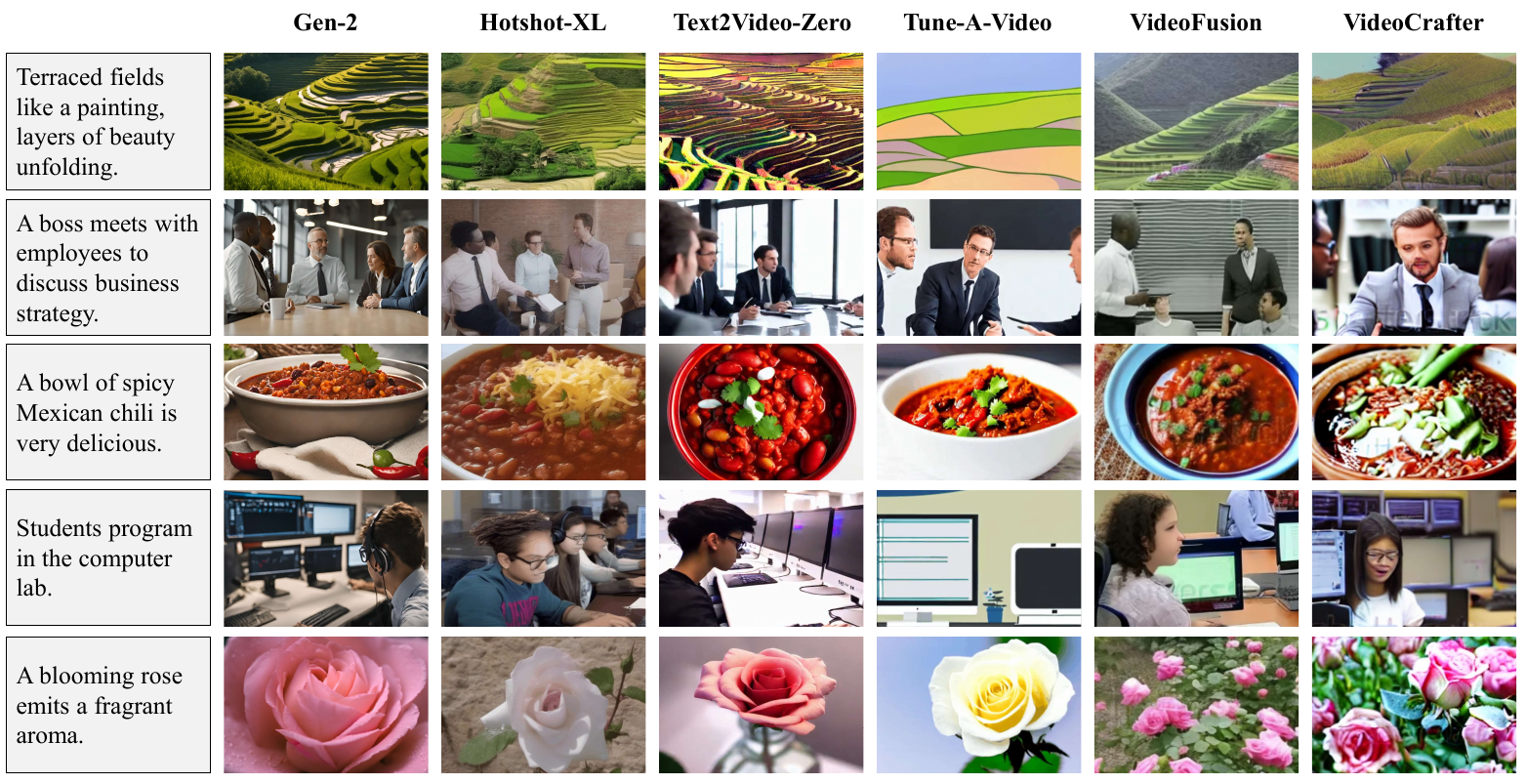}
    \caption{Some example frames in proposed LGVQ dataset.}
    \label{fig_dataset}
    \vspace{-0.5cm}
\end{figure*}

In total, LGVQ contains $2,808$ AIGC videos generated by $6$ T2V methods using $468$ text prompts. Screenshots of selected video frames, presented in  Fig.~\ref{fig_dataset}, provide a straightforward representation of the generated videos. 


\subsection{Subjective Quality Assessment Experiment}

We conducted a subjective quality assessment on the LGVQ dataset to derive quality labels for each AIGC video. Specifically, we consider three critical quality dimensions: (1) \textbf{Spatial quality}, which examines the visual appearance of individual frames; (2) \textbf{Temporal quality}, which evaluates the coherence across video frames; (3) \textbf{Text-video alignment}, which measures the correspondence between the video content and the accompanying text prompt. 

\begin{figure}
    \centering
    \includegraphics[width=0.7\linewidth]{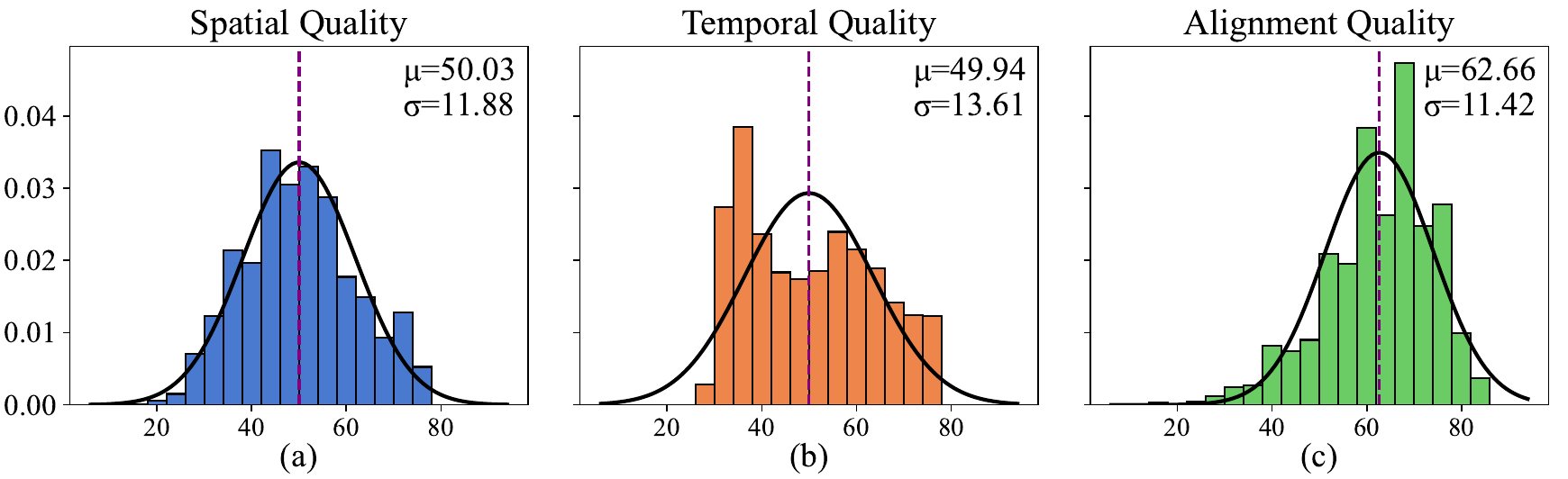}
    \caption{Illustration of the MOSs probability density of the LGVQ dataset.}
    \label{fig_probability}
\end{figure}


\subsubsection{Experimental Methodology and Configuration}
\begin{itemize}
\item \textbf{Participants:} $60$ participants were required to assess the video quality. The age range of the participants was between $20$ and $30$ years old, consisting of $34$ males and $26$ females. All participants had normal or corrected-to-normal vision.
\item \textbf{Test Condition:} The experiments were conducted in a controlled environment to minimize external variables that could influence the judgments of the participants. The device included 27-inch display monitors with resolution of 4K. The viewing distance was set at 70 cm.  The room lighting was maintained at a consistent level of 300 lux to ensure uniformity across all viewing sessions.
\item \textbf{Quality Rating:} We adopt the single-stimulus rating method, during the subjective experiments, participants rated each dimension on a scale from $1$ to $5$, where $1$ represents the lowest quality and $5$ is the highest. We listed the specific rating criteria in the Supplemental File.
\end{itemize}

Before the formal assessments, participants underwent a training session where they reviewed sample videos that were not included in the formal experiment. This session was intended to familiarize them with the evaluation criteria and the rating interface. Example distortions and quality attributes were discussed to calibrate participant expectations and rating standards. 

In the formal experiment, the resolution of the short side of all videos is set to $768$ pixels, while maintaining the original aspect ratio of each video. The $2,808$ videos were divided into six groups, each containing $468$ videos covering all $468$ prompts, and there was no overlapping between participants across different groups.  In each group, each video was generated by one of the six T2V models and rated by ten different subjects.  To avoid visual fatigue, each session lasted no longer than $30$ minutes, ensuring participants could maintain a high level of attention and accuracy in their ratings.

\subsection{Data Processing and Analysis}
After the subjective experiment, we collected the perceptual ratings and conducted data analysis. We follow the recommended method in ~\cite{Methodology} to process the subjective ratings collected during the experiment. Outlier ratings are detected and removed if they deviate by more than \(2\sigma\) (for normal distributions) or \(\sqrt{20}\sigma\) (for non-normal distributions) from the mean rating for that condition. Observers contributing more than 5\% of outlier ratings are excluded from the analysis.
For each test condition, the mean score (\(\mu_i\)) and standard deviation (\(\sigma_i\)) are calculated based on all valid ratings provided by observer \(i\).
To mitigate individual bias, each raw score \(s_{ij}\) is normalized to a Z-score.
Finally, the \textbf{Mean Opinion Score (MOS)} for each test condition \(j\) is computed as the average of the normalized Z-scores across all observers (\(M_j\)), Z-score $Z_{ij}$ is linearly rescaled to lie in the range of [0, 100].

\subsection{MOS Distribution Analysis}
The MOSs distributions of the three dimensions are illustrated in Fig.~\ref{fig_probability}, showing that all three follow an approximately Gaussian distribution. Notably, \textbf{the text-video alignment distributions exhibit higher central values compared to the temporal quality distribution}, suggesting that existing T2V models perform better in text-video alignment than in spatial and temporal quality. Additionally, the variance is small in the spatial quality and text-video alignment dimension, indicating a more consistent quality in these aspect compared to the temporal dimensions.



\subsubsection{MOS Analysis for T2V models}

We calculate the average MOSs of six T2V models across three quality dimensions to analyze their subjective performance in generating videos. The results are shown in Fig.~\ref{fig_all_model_comparison}.
Gen-2 outperforms all other models across all three dimensions, showing a particularly significant margin in the spatial and temporal quality, which highlights its better capability in generating visually high-quality content. 
Following Gen-2, Hotshot-XL, VideoFusion, and VideoCrafter perform similarly, with their performance in text-video alignment surpassing temporal quality, and both exceeding their spatial quality.
Text2Video-Zero achieves comparable results to these three T2V models in spatial quality and text-video alignment but falls significantly behind in temporal quality, revealing its weaker ability to produce temporally consistent video content. 
Tune-A-Video performs the worst across all three quality dimensions, with its alignment quality outperforming its spatial quality, while both exceed its temporal quality.

\subsubsection{MOS Analysis for Text Prompts}


\begin{figure}
    \centering
    \begin{minipage}{0.58\textwidth}
        \centering
        \includegraphics[width=1\linewidth]{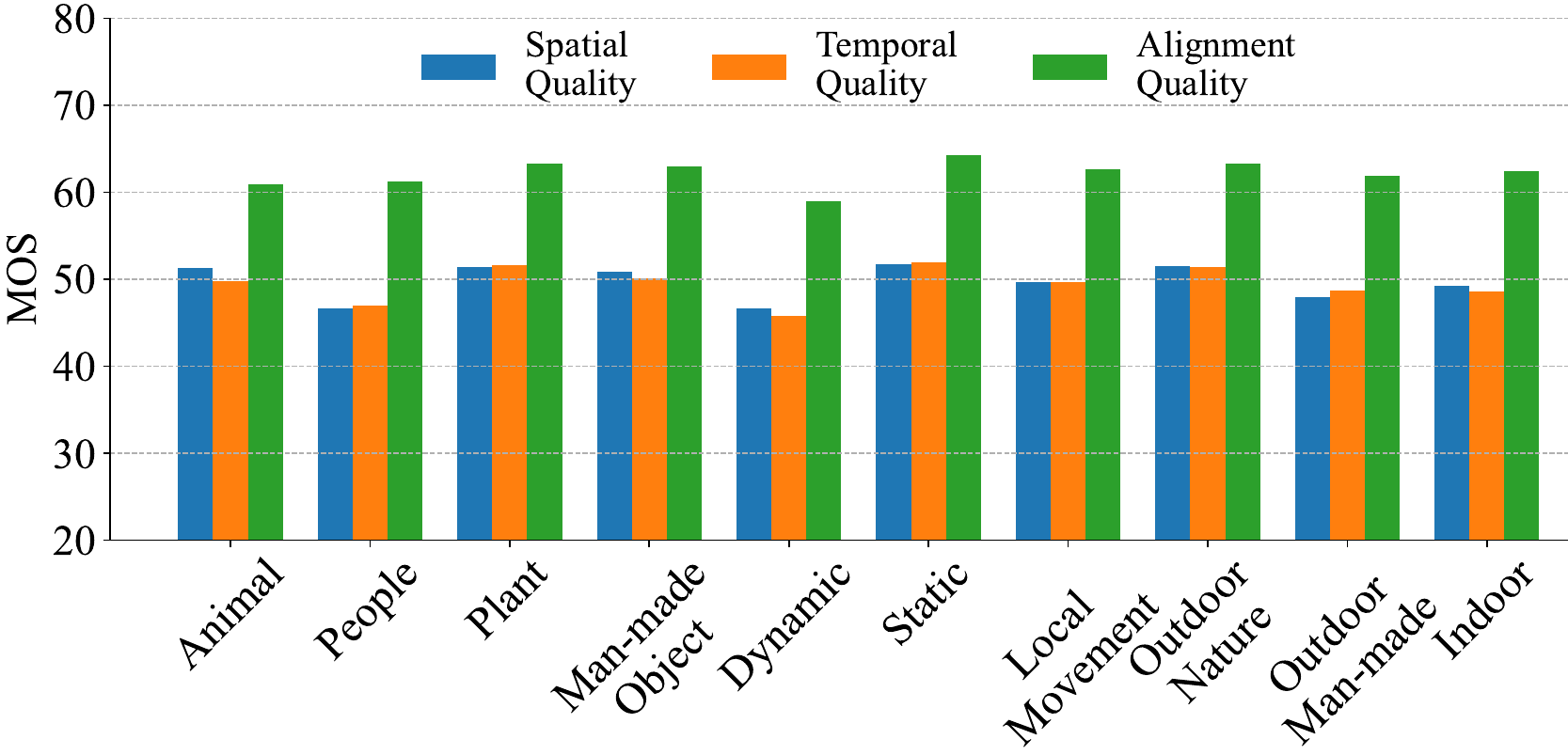} 
        \caption{Comparison of MOSs of different generation elements. The quality scores are adjusted to range from 20 to 80.}
        \label{fig_all_cls_comparison}
    \end{minipage}
    \hfill
    \begin{minipage}{0.4\textwidth}
        \centering
        \includegraphics[width=1\linewidth]{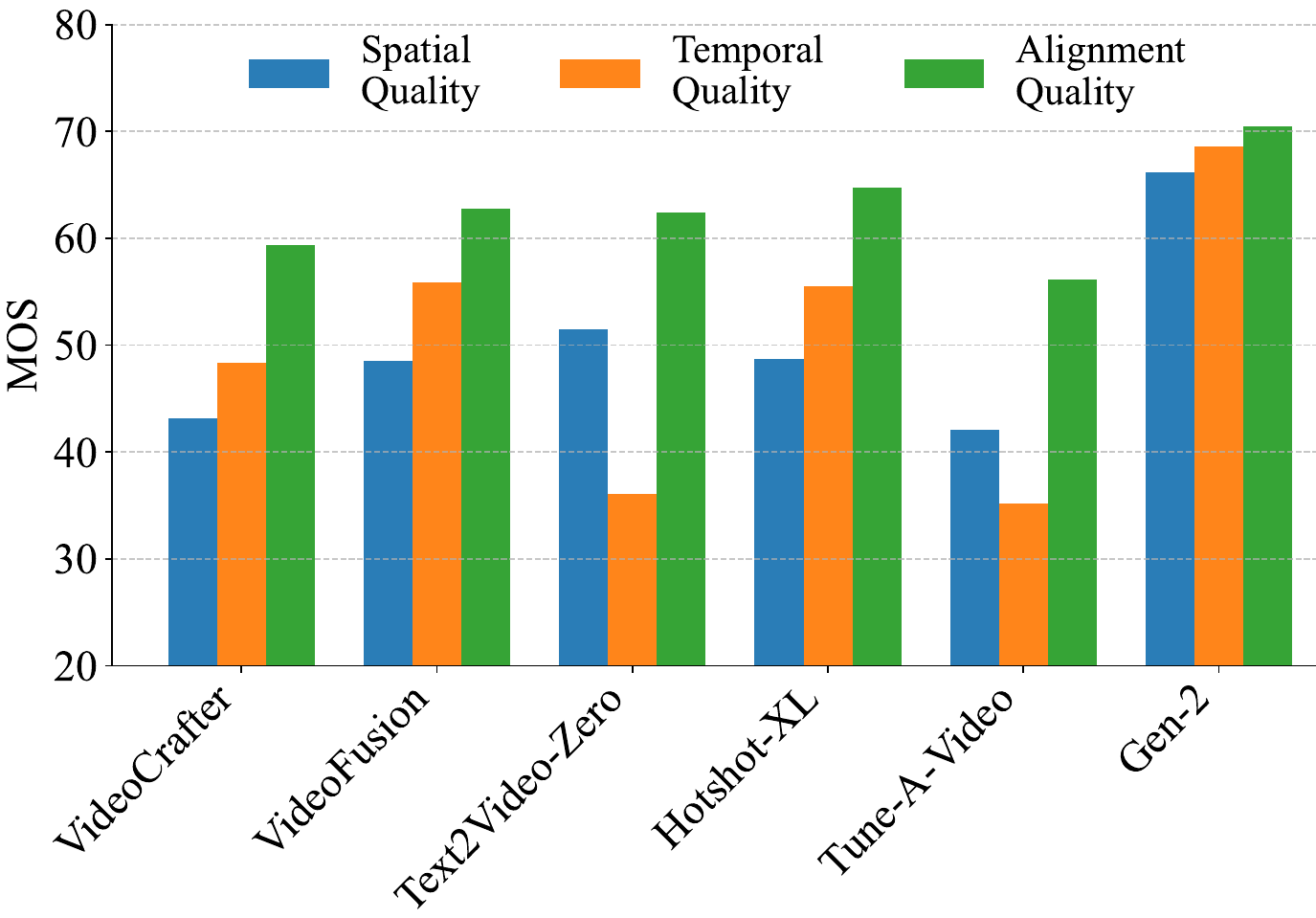} 
        \caption{Comparison of MOSs of different generation models. }
        \label{fig_all_model_comparison}
    \end{minipage}
\end{figure}



Fig.~\ref{fig_all_cls_comparison} shows the average MOSs for $3$ text prompt categories and their $10$ subcategories. 
We observe that the text-video alignment quality remains relatively stable across all text prompt categories, which indicates that current T2V models can generate corresponding video content aligned with text to some extent, without significant bias toward specific semantic categories. For the foreground categories, the spatial and temporal quality of the `human' category is significantly lower than that of other categories, likely due to the complexity of human actions, postures, and expressions. Regarding motion states, static states outperform local movement states in spatial quality, temporal quality, and text-video alignment, while dynamic states perform the worst. This is reasonable, as large-magnitude motion is inherently more challenging to generate. For background content, outdoor man-made scenes and indoor scenes perform comparably, while outdoor natural scenes achieve better scores. This may be because man-made content typically has more complex structures, making it more difficult to generate accurately.

\section{AIGC VQA Benchmark}
\label{sec_benchmark}
In this section, we benchmark existing quality metrics to validate their effectiveness in assessing the quality of AIGC videos. Focusing on three quality dimensions: \textbf{spatial quality}, \textbf{temporal quality}, and \textbf{text-to-video alignment}, we select three corresponding types of quality metrics: IQA metrics to evaluate the spatial quality, VQA metrics to assess the temporal quality, and CLIP-based metrics to measure the text-to-video alignment. Through establishing the comprehensive AIGC-VQA benchmark, we can diagnose the strengths and weakness of current quality metrics in evaluating AIGC videos, shedding light on designing effective AIGC-specific quality metrics.

\begin{table*}[t]
\centering
  \caption{The benchmark performance of existing IQA, VQA, and text-visual alignment methods on the LGVQ dataset is presented. NSC and AIGC refer to natural scene content and AI-generated content, respectively. The best-performing metric is highlighted in bold, while the second-best metric is underlined.}
  \label{tab_benchmark}
  \resizebox{0.95\textwidth}{!}{
  \begin{tabular}{clcccccccccc}
    \toprule
                               & \multirow{2}{*}{Method}                        & \multirow{2}{*}{\makecell[l]{Pre-training /\\Initialization}}  & \multirow{2}{*}{\makecell[l]{Model\\Type}}  & \multicolumn{3}{c}{Video-level}   & \multicolumn{3}{c}{Model-level}        \\
                                \cmidrule(r){5-7}  \cmidrule(r){8-10}                                                                                                                                                                                                               
                               &                                                &                                                                &                     & SRCC               & KRCC              & PLCC              & SRCC              & KRCC              & PLCC          \\
    \midrule                                                                                                                                                                                                                                                                                  
    \multirow{14}{*}{Spatial}  & NIQE (ISPL, 2012)~\cite{NIQE}                  & NA (\textit{handcraft})                                        & NSC                 & 0.228              & 0.127             & 0.293             & 0.425             & 0.333             & 0.624         \\
                               & BRISQUE (TIP, 2012)~\cite{BRISQUE}             & NA (\textit{handcraft})                                        & NSC                 & 0.255              & 0.155             & 0.357             & 0.434             & 0.354             & 0.672         \\
                               & IS (NIPS, 2016)~\cite{IS}                      & NA (\textit{handcraft})                                        & AIGC                & 0.184              & 0.119             & 0.228             & 0.418             & 0.326             & 0.653         \\
                               & FID (NIPS, 2017)~\cite{FID}                    & NA (\textit{handcraft})                                        & AIGC                & 0.197              & 0.137             & 0.247             & 0.451             & 0.372             & 0.697         \\
                               & HyperIQA (CVPR, 2020)~\cite{HyperIQA}          & KonIQ-10k~\cite{KonIQ_10k}                                     & NSC                 & 0.296              & 0.217             & 0.376             & 0.502             & 0.394             & 0.715         \\
                               & UNIQUE (TIP, 2021)~\cite{UNIQUE}               & KonIQ-10K~\cite{KonIQ_10k}                                     & NSC                 & 0.130              & 0.089             & 0.231             & 0.468             & 0.395             & 0.700         \\
                               & MUSIQ (ICCV, 2021)~\cite{MUSIQ}                & KonIQ-10K~\cite{KonIQ_10k}                                     & NSC                 & \underline{0.389}  & 0.221             & \underline{0.403} & 0.526             & \underline{0.417} & 0.756         \\
                               & TReS (WACV, 2022)~\cite{TReS}                  & KonIQ-10k~\cite{KonIQ_10k}                                     & NSC                 & 0.325              & 0.202             & 0.377             & 0.504             & 0.398             & 0.730         \\
                               & StairIQA (JSTSP, 2023)~\cite{StairIQA}         & KonIQ-10k~\cite{KonIQ_10k}                                     & NSC                 & 0.334              & 0.209             & 0.393             & 0.517             & 0.406             & 0.746         \\
                               & CLIP-IQA+ (AAAI, 2023)~\cite{CLIP_IQA}         & KonIQ-10K~\cite{KonIQ_10k}                                     & NSC                 & 0.341              & 0.221             & 0.355             & 0.519             & 0.412             & 0.743         \\
                               & LIQE (CVPR, 2023)~\cite{LIQE}                  & KonIQ-10k~\cite{KonIQ_10k}                                     & NSC                 & 0.174              & 0.116             & 0.209             & 0.467             & 0.364             & 0.715         \\
                               & TOPIQ (TIP, 2024)~\cite{TOPIQ}                 & KonIQ-10K~\cite{KonIQ_10k}                                     & NSC                 & 0.347              & 0.227             & 0.389             & 0.534             & 0.382             & 0.769         \\
                               & Q-Align (ICML, 2024)~\cite{qalign}             & \textit{fused} ~\cite{KonIQ_10k},~\cite{SPAQ},~\cite{PatchVQ},~\cite{Ava} & NSC      & 0.381              & \underline{0.239} & 0.401             & \underline{0.542} & 0.375             & \underline{0.782}\\
                               & MA-AGIQA (ACMMM, 2024)~\cite{MA_AGIQA}         & AGIQA-3k~\cite{AGIQA_3K}                                       & AIGC                & \textbf{0.402}     & \textbf{0.247}    & \textbf{0.449}    & \textbf{0.561}    & \textbf{0.454}    & \textbf{0.797}\\
    \midrule                                                                                                                                                                                                                                                                                  
    \multirow{16}{*}{Temporal} & FVD (Arxiv, 2018)~\cite{FVD}                   & NA (\textit{handcraft})                                        & AIGC                & 0.213              & 0.149             & 0.385             & 0.416             & 0.338             & 0.583         \\
                               & KVD (Arxiv, 2018)~\cite{KVD}                   & NA (\textit{handcraft})                                        & AIGC                & 0.226              & 0.153             & 0.393             & 0.428             & 0.346             & 0.597         \\
                               & TLVQM (TIP, 2019)~\cite{TLVQM}                 & NA (\textit{handcraft})                                        & NSC                 & 0.286              & 0.189             & 0.437             & 0.471             & 0.384             & 0.629         \\
                               & RAPIQUE (JSP, 2021)~\cite{RAPIQUE}             & NA (\textit{handcraft})                                        & NSC                 & 0.313              & 0.212             & 0.451             & 0.501             & 0.401             & 0.707         \\
                               & VSFA (ACMMM, 2019)~\cite{VSFA}                 & KoNViD-1k~\cite{KoNViD_1k}                                     & NSC                 & 0.295              & 0.192             & 0.451             & 0.467             & 0.370             & 0.623         \\
                               & VIDEAL (TIP, 2021)~\cite{VIDEAL}               & KoNViD-1k~\cite{KoNViD_1k}                                     & NSC                 & 0.238              & 0.155             & 0.421             & 0.448             & 0.352             & 0.638         \\
                               & PatchVQ (CVPR, 2021)~\cite{PatchVQ}            & LSVQ~\cite{PatchVQ}                                            & NSC                 & 0.275              & 0.181             & 0.439             & 0.516             & 0.415             & 0.689         \\
                               & SimpleVQA (ACMMM, 2022)~\cite{simpleVQA}       & LSVQ~\cite{PatchVQ}                                            & NSC                 & 0.271              & 0.182             & 0.419             & 0.507             & 0.406             & 0.684         \\
                               & FastVQA (ECCV, 2023)~\cite{FAST_VQA}           & LSVQ~\cite{PatchVQ}                                            & NSC                 & 0.374              & 0.255             & 0.473             & 0.586             & 0.452             & \underline{0.828}\\
                               & DOVER (ICCV, 2023)~\cite{dover}                & LSVQ~\cite{PatchVQ}                                            & NSC                 & 0.254              & 0.164             & 0.514             & 0.498             & 0.384             & 0.702         \\
                               & LMM-VQA (Arxiv, 2024)~\cite{LMM_VQA}           & \textit{fused}~\cite{PatchVQ,WIT}                              & NSC                 & 0.374              & \underline{0.260} & 0.494             & 0.486             & 0.333             & \textbf{0.838}\\
                               & T2VQA (ACMMM, 2024)~\cite{PatchVQ}             & LSVQ~\cite{PatchVQ}                                            & AIGC                & \textbf{0.394}     & \textbf{0.267}    & \textbf{0.519}    & \textbf{0.604}    & \textbf{0.468}    & 0.837         \\
                               & Motion Smoothness (CVPR, 2024)~\cite{VBench}   & AMT~\cite{Amt}                                                 & AIGC                & 0.299              & 0.189             & 0.441             & 0.527             & 0.416             & 0.748         \\
                               & Temporal Flickering (CVPR, 2024)~\cite{VBench} & RAFT~\cite{RAFT}                                               & AIGC                & 0.372              & 0.256             & 0.493             & 0.576             & 0.450             & 0.794         \\
                               & Action-Score (CVPR, 2024)~\cite{EvalCrafter}   & VideoMAE V2~\cite{Videomae_v2}                                 & AIGC                & 0.316              & 0.224             & 0.454             & 0.542             & 0.439             & 0.765         \\
                               & Flow-Score (CVPR, 2024)~\cite{EvalCrafter}     & RAFT~\cite{RAFT}                                               & AIGC                & \underline{0.387}  & 0.258             & \underline{0.509} & \underline{0.593} & \underline{0.461} & 0.825         \\
    \midrule                                                                                                                                                                                                                                                                               
    \multirow{9}{*}{Alignment} & CLIP (ICML, 2021)~\cite{CLIP}                  & WIT~\cite{WIT}                                                 & NSC                 & 0.324              & 0.239             & 0.388             & 0.628             & 0.533             & 0.718         \\
                               & BLIP (ICML, 2022)~\cite{BLIP}                  & \textit{fused}~\cite{Flickr30K,MSCOCO}                         & NSC                 & 0.379              & 0.260             & 0.389             & 0.641             & 0.537             & 0.729         \\
                               & viCLIP (ArXiv, 2022)~\cite{viCLIP}             & InternVid-10M~\cite{Internvid}                                 & AIGC                & 0.397              & 0.280             & 0.421             & 0.725             & 0.589             & 0.761         \\
                               & ImageReward (NIPS, 2023)~\cite{ImageReward}    & ImageRewardDB~\cite{ImageReward}                               & AIGC                & 0.369              & 0.255             & 0.371             & 0.836             & 0.690             & 0.877         \\
                               & PickScore (NIPS, 2023)~\cite{PickScore}        & Pick-a-Pic-v2~\cite{PickScore}                                 & AIGC                & 0.381              & 0.262             & 0.382             & \underline{0.885} & \textbf{0.733}    & \underline{0.924} \\
                               & HPSv1 (ICCV, 2023)~\cite{HPSv1}                & HPDv1~\cite{HPSv1}                                             & AIGC                & 0.248              & 0.171             & 0.339             & 0.733             & 0.602             & 0.785         \\
                               & HPSv2 (ArXiv, 2023)~\cite{HPSv2}               & HPDv2~\cite{HPSv2}                                             & AIGC                & 0.325              & 0.223             & 0.395             & 0.798             & 0.641             & 0.832         \\
                               & GenEval (NIPS, 2024)~\cite{Geneval}            & \textit{fused}~\cite{Flickr30K,MSCOCO}                         & AIGC                & \underline{0.412}  & \underline{0.297} & \underline{0.448} & 0.859             & \underline{0.704} & 0.893         \\
                               & VQAScore (ECCV, 2024)~\cite{VQAScore}          & \textit{fused}~\cite{VQAv2,GQA,LAION_5B}                       & AIGC                & \textbf{0.461}     & \textbf{0.335}    & \textbf{0.493}    & \textbf{0.887}    & \textbf{0.733}    & \textbf{0.931}\\
    \bottomrule          
    \end{tabular}
  }
\end{table*}

\subsection{Compared Quality Metrics} 
\begin{itemize}
\item \textbf{Spatial Quality Metrics.} We utilize the IQA metrics to evaluate the spatial quality of AIGC videos. Specifically, we select four knowledge-driven IQA metrics, including two NSS-based methods, NIQE~\cite{NIQE} and BRISQUE~\cite{BRISQUE}, and two deep features-based methods, Inception Score (IS)~\cite{IS}, and FID~\cite{FID}. Additionally, we choose ten data-driven IQA metrics: UNIQUE~\cite{UNIQUE}, HyperIQA~\cite{HyperIQA}, MUSIQ~\cite{MUSIQ}, StairIQA~\cite{StairIQA}, TReS~\cite{TReS}, TOPIQ~\cite{TOPIQ}, CLIP-IQA~\cite{CLIP_IQA}, LIQE~\cite{LIQE}, Q-Align~\cite{qalign}, MA-AGIQA~\cite{MA_AGIQA}. Among these metrics, IS, FID, and MA-AGIQA are specifically designed for evaluating AIGC images.
\item \textbf{Temporal Quality Metrics.} We utilize VQA metrics to evaluate the temporal quality. Specifically, we choose four knowledge-driven metrics, including TLVQM~\cite{TLVQM}, RAPIQUE~\cite{RAPIQUE}, FVD~\cite{FVD}, KVD~\cite{KVD}, seven data-driven metrics, including VSFA~\cite{VSFA}, PatchVQ~\cite{PatchVQ}, SimpleVQA~\cite{simpleVQA}, FastVQA~\cite{FAST_VQA}, DOVER~\cite{dover}, T2VQA~\cite{T2VQA}, and LMM-VQA~\cite{LMM_VQA}, and four AIGC-specific temporal descriptors: Motion Smoothness~\cite{VBench}, Temporal Flickering~\cite{VBench}, Action-Score~\cite{EvalCrafter}, and Flow-Score~\cite{EvalCrafter}. Among these metrics, FVD, KVD, T2VQA, Motion Smoothness, Temporal Flickering, Action-Score, and Flow-Score are specifically designed for evaluating AIGC videos.
\item \textbf{Text-video Alignment.} We select two types of text and visual content alignment metrics to evaluate text-video alignment: CLIP-based metrics and visual question-answering-based metrics. The CLIP-based metrics include CLIP~\cite{CLIPScore}, BLIP~\cite{BLIP}, viCLIP~\cite{viCLIP}, ImageReward~\cite{ImageReward}, PickScore~\cite{PickScore}, HPSv1~\cite{HPSv1}, and HPSv2~\cite{HPSv2}, and the visual question answering-based metrics include GenEval~\cite{Geneval}, and VQAScore~\cite{VQAScore}.
\end{itemize}

We evaluate the compared quality metrics on the whole LGVQ dataset in a zero-shot setting. For the data-driven quality metrics, the corresponding training datasets are provided in Table~\ref{tab_benchmark}. The performance of these metrics is measured using three standard indices: \textbf{Spearman’s rank correlation coefficient (SRCC)}, \textbf{Kendall’s rank correlation coefficient (KRCC)}, and \textbf{Pearson’s linear correlation coefficient (PLCC)}, with higher values indicating stronger alignment with human perception. The evaluation is performed at two levels: \textbf{the video-level}, which assesses the ability of the quality metrics to evaluate individual AIGC videos, and \textbf{the model-level}, which examines their capability to assess the generation performance of specific T2V models.

\vspace{0.2cm}
\subsection{Performance Analysis} 

The benchmarking results are listed in Table~\ref{tab_benchmark}, which show that nearly all quality metrics perform worse when evaluating the quality of AIGC videos. A detailed analysis of the performance across three quality dimensions is presented as follows:
\begin{itemize}
\item \textbf{Spatial Quality Assessment Metrics:} Knowledge-driven IQA metrics, including NSC-based methods (\textit{e.g.}, NIQE, BRISQUE) and AIGC-specific metrics (\textit{e.g.}, IS, FID), exhibit the poorest performance when assessing the spatial quality of AIGC videos. In contrast, data-driven IQA metrics generally outperform their knowledge-driven counterparts. Notably, \textbf{MA-AIGQA}, a DNN-based IQA method trained on the AIGC IQA dataset, achieves the best performance among all the compared metrics. This underscores the effectiveness of training quality assessment models on AIGC-specific datasets, which enables them to better capture distortions unique to AIGC content. 

\item \textbf{Temporal Quality Assessment Metrics:} A similar trend is observed for temporal quality metrics. Knowledge-driven VQA methods (e.g., TLVQM, RAPIQUE), which are effective for assessing natural scene video quality, perform poorly when evaluating the temporal quality of AIGC videos. Conversely, data-driven models demonstrate better adaptability to generated content. Furthermore, AIGC-specialized metrics, such as \textbf{T2VQA}, \textbf{Temporal Flickering}, and \textbf{Motion Smoothness}, which are explicitly designed for AIGC tasks, achieve stronger correlations with human evaluations. This highlights the necessity of tailoring temporal quality assessment methodologies to better align with the unique characteristics of AIGC videos.

\item \textbf{Text-video Alignment Metrics:} Methods fine-tuned on AIGC alignment datasets, such as \textbf{ImageReward} and \textbf{PickScore}, slightly outperform their CLIP-based counterparts (\textit{e.g.}, CLIP, BLIP), demonstrating that fine-tuning on AIGC-specific datasets enhances the ability to evaluate text-video alignment. Additionally, the video-based CLIP method \textbf{viCLIP} achieves the best performance among CLIP-based approaches, highlighting the importance of incorporating temporal-related features for effective text-video alignment. Furthermore, we observe that visual question-answering-based methods, such as \textbf{GenEval} and \textbf{VQAScore}, significantly outperform CLIP-based methods. This may be attributed to the fact that visual question-answering-based methods leverage the strong generalization capabilities of visual question-answering models, enabling better performance across diverse AIGC datasets.

\item The results show that \textbf{model-level evaluation} consistently outperforms \textbf{video-level evaluation}, suggesting that assessing the quality of individual videos is inherently more challenging than evaluating the overall performance of T2V models. 
\end{itemize}

In summary, we conclude that existing quality metrics fail to accurately evaluate the quality of AIGC videos, highlighting the urgent need for more effective and specialized quality assessment models tailored to AIGC video content.

\begin{figure*}
    \centering
    \includegraphics[width=0.95\linewidth]{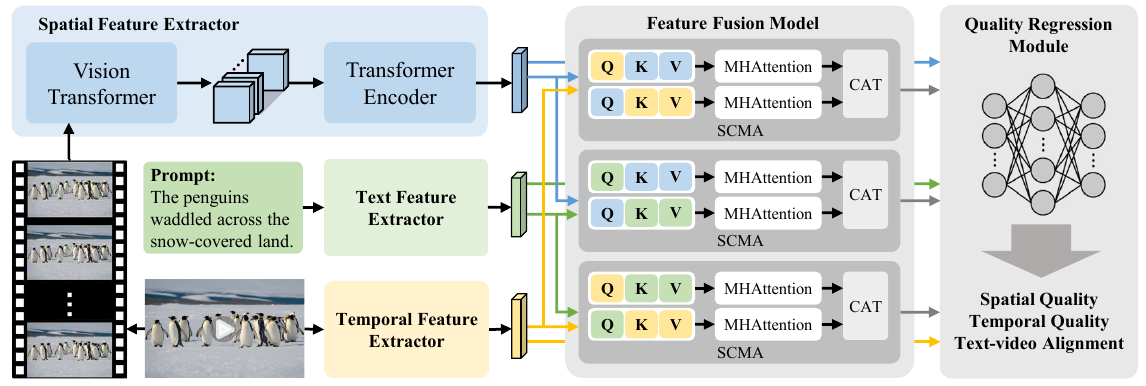}
    \caption{The framework of the UGVQ model. The spatial, temporal, and text feature extractors are utilized to extract features from key frames, entire videos, and text prompts, respectively. The feature fusion module integrates these spatial, temporal, and text features using symmetric cross-modality attention layers to generate a fused feature representation. The quality regressor module maps the fused features into the spatial quality, temporal quality, and text-video alignment scores.}
    \label{fig_UGVQ}
\end{figure*}

\section{Proposed Method}
\label{sec_method}

\subsection{Preliminaries}
Given a video $\bm{x}= \{\bm x_i\}_{i=0}^{N-1}$ generated from a text prompt $p$, where $\bm x_i\in\mathbb{R}^{H\times W \times 3}$ denotes the $i$-th frame. Here, $H$ and $W$ are the height and width of each frame, and $N$ is the total number of frames. The objective of the UGVQ metric is to compute the spatial quality, temporal quality, and text-video alignment of $\bm{x}$ with respect to $p$. We formulate it as:
\begin{align}
    \hat{q_s}, \hat{q_t}, \hat{q_a} = {\mathbf{UGVQ}}({\bm x}, p),
\end{align}
where $\hat{q_s}, \hat{q_t}, \hat{q_a}$ represent spatial quality, temporal quality, and text-video alignment scores.

To evaluate the spatial quality, temporal quality, and text-video alignment, a straightforward strategy is to individually extract spatial, temporal, and text-related features, and then use these features to compute the corresponding scores. Following this intuitive methodology, we decompose $\rm UGVQ$ into five components: a spatial feature extractor, a temporal feature extractor, a text feature extractor, a feature fusion module, and a quality regressor, as shown in Fig.~\ref{fig_UGVQ}. These components are described in detail below.

\subsection{Spatial Feature Extractor}
Spatial quality focuses on the visual fidelity of video frames. Therefore, the proposed spatial feature extractor is designed to capture quality-aware spatial features at the frame level. Given the significant spatial redundancy between consecutive frames in a video, we first perform temporal downsampling on the video sequence, converting $\bm x=[\bm x_0,\bm x_1, \ldots, \bm x_{N-1}]$, into a lower frame rate sequence, $\bm y=[\bm y_0,\bm y_1, \ldots, \bm y_{N_s-1}]$, where $\bm{y}_i = \bm{x}_{\lfloor N/N_s\times i \rfloor}$, and $N_s$ is the number of frames in $\bm{y}$.

Previous VQA models typically employ pre-trained IQA models as spatial feature extractors, enabling efficient extraction of quality-related features. However, as discussed in Section~\ref{sec_benchmark}, IQA models trained on NSC IQA datasets exhibit suboptimal performance when assessing the spatial quality of AIGC videos. To address this limitation, we employ a ViT pre-trained on an AIGC IQA dataset (\textit{i.e.,} Pick-a-pic dataset~\cite{PickScore}) as the spatial feature extractor, specifically designed to capture AIGC-specific distortions. Subsequently, we extract the spatial features for each frame in $\bm y$:

\begin{align}
    F_{\text{spatial},i} = {\mathbf{ViT}}({\bm y}_i),
\end{align}
where $\mathbf{ViT}$ denotes the pre-trained ViT model and $F_{\text{spatial},i}$ represents the spatial features of video frame ${\bm y}_i$.

Since the human visual system does not perceive video quality as being equally influenced by the quality of each individual frame, we utilize a sequence modeling network (\textit{i.e.,} Transformer encoder) to capture the non-linear relationships between video-level features and frame-level features:
\begin{align}
    F_{\text{spatial}} = {\mathbf{Transformer}}([F_{\text{spatial},i}]_{i=0}^{N_s-1}),
\end{align}
where $\mathbf{Transformer}$ denotes the Transformer encoder network and $F_{\text{spatial}}$ represents the video-level spatial features of video ${\bm x}$.

\subsection{Temporal Feature Extractor}
Motion features play a crucial role in identifying temporal quality degradation caused by discontinuous actions and frame flicker. Action recognition networks, such as SlowFast~\cite{SlowFast}, are widely used to model features related to action detection and classification, as they effectively capture patterns of motion and frame discontinuity of distorted videos. Consequently, we employ the pre-trained action recognition network, SlowFast~\cite{SlowFast}, as a temporal feature extractor to derive temporal-related features for AIGC videos. Specifically, given a video $\bm x$ and the action recognition network $\mathbf{SlowFast}$, we extract temporal features as follows:

\begin{equation} \label{eq_t_v} F_{\text{temporal}} = \mathbf{SlowFast}({\bm x}), \end{equation}
where $\mathbf{SlowFast}$ denotes the SlowFast action recognition network, and $F_{\text{temporal}}$ represents the temporal features of the video $\bm x$.

\subsection{Text Feature Extractor}
For AIGC videos, the alignment between the prompt and the generated video content is a critical aspect of quality assessment. Extracting semantic content features from the prompt is essential for evaluating this alignment. As discussed in Session~\ref{sec_benchmark}, CLIP-based methods show a notable ability to assess the alignment between textual and visual content while maintaining a simplistic structure. Therefore, we adopt CLIP to extract the semantic-aware text features of the prompt.
For a given prompt $p$, we can extract the text feature through the text encoder of CLIP:
\begin{equation}
  \label{eq_sem_p}
    F_{text} = \mathbf{CLIP}_\mathbf{Text}(p),
\end{equation}
where $F_{text}$ represent the text feature of prompt $p$ and $\mathbf{CLIP}_\mathbf{Text}$ is the text encoder of CLIP.

\subsection{Feature Fusion Module}

After extracting spatial, temporal, and textual features, we introduce a feature fusion module to effectively capture the relationships among these multi-modal features and generate unified quality-aware features for quality regression. The proposed feature fusion module consists of three \textbf{symmetric cross-modality attention (SCMA)} layers, each responsible for capturing interactions between different modality pairs. For instance, given two modality features, $F_{\text{spatial}}$ and $F_{\text{text}}$, the SCMA layer first uses $F_{\text{spatial}}$ as the query and $F_{\text{text}}$ as the key and value, employing the multi-head attention mechanism to derive the fused features $F_{\text{spatial},\text{text}}$. Next, by treating $F_{\text{text}}$ as the query and $F_{\text{spatial}}$ as the key and value, the SCMA layer produces the fused features $F_{\text{text},\text{spatial}}$. Finally, $F_{\text{spatial},\text{text}}$ and $F_{\text{text},\text{spatial}}$ are concatenated to produce the fused features from the SCMA layer:
\begin{equation}
\begin{aligned}
  \label{eq_sem_p}
    F_{\text{spatial},\text{text}} &= \textbf{MHAttention}(F_{\text{spatial}},F_{\text{text}}, F_{\text{text}}), \\
    F_{\text{text},\text{spatial}} &= \textbf{MHAttention}(F_{\text{text}},F_{\text{spatial}}, F_{\text{spatial}}), \\
    F_{\text{fuse},1} &= \textbf{CAT}(F_{\text{spatial},\text{text}}, F_{\text{text},\text{spatial}}),
\end{aligned}
\end{equation}
where $\textbf{MHAttention}(.)$ represents the multi-head attention operator, with its first, second, and third parameters corresponding to the inputs for the query, key, and value, respectively, and $\textbf{CAT}$ represents the concatenation operator. We summary this procedure as:
\begin{equation}
  \label{eq_sem_p}
    F_{\text{fuse},1} = \textbf{SCMA}(F_{\text{spatial}}, F_{\text{test}}),
\end{equation}
where $\textbf{SCMA}$ refers to the symmetric cross-modality attention layer.

Similarly, the SCMA layer is applied to other modality pairs including $F_{\text{spatial}}$ and $F_{\text{temporal}}$ as well as $F_{\text{temporal}}$ and $F_{\text{text}}$ to derive the fused features:
\begin{equation}
\begin{aligned}
    F_{\text{fuse},2} &= \textbf{SCMA}(F_{\text{spatial}}, F_{\text{temporal}}),\\
    F_{\text{fuse},3} &= \textbf{SCMA}(F_{\text{temporal}}, F_{\text{text}}).
\end{aligned}
\end{equation}

Finally, we concatenate the fused features $F_{\text{fuse},1}$, $F_{\text{fuse},2}$, and $F_{\text{fuse},3}$ with the original single-modality features $F_{\text{spatial}}$, $F_{\text{temporal}}$, and $F_{\text{text}}$ to form the unified quality-aware features $F_{q}$:
\begin{equation}
  \label{eq_sem_p}
    F_{q} = \textbf{CAT}(F_{\text{spatial}}, F_{\text{temporal}}, F_{\text{text}}, F_{\text{fuse},1}, F_{\text{fuse},2}, F_{\text{fuse},3}).
\end{equation}

\subsection{Quality Regressor}
The quality regressor is designed to map the quality-aware features into multi-dimensional quality scores. For simplicity, we employ a \textbf{multi-layer perceptron (MLP)} as the quality regressor to output spatial quality, temporal quality, and text-video alignment scores:
\begin{equation}
    \begin{split}
        \label{eq_reg}
            \hat{q_s}, \hat{q_t}, \hat{q_a} = \textbf{MLP}( F_q ), 
    \end{split}
\end{equation}
where $\textbf{MLP}$ represents the quality regression module, and $\hat{q_s}, \hat{q_t}, \hat{q_a}$ correspond to the derived spatial quality, temporal quality, and text-video alignment scores, respectively.
The loss function used to optimize the UGVQ models consists of the \textbf{mean absolute error (MAE)} loss and rank loss~\cite{rank_loss}. The MAE loss is used to make the evaluated quality scores close to the ground truth, and rank loss makes the model better distinguish the relative quality of videos.

\section{Experiments}
\label{sec_experiments}
\subsection{Experiment Settings}

\begin{table*}[]
\caption{The video-level performance of the proposed UGVQ metric and the compared quality metrics on the LGVQ, FETV, and MQT datasets. The best-performing metric is highlighted in bold, while the second-best metric is underlined.}
\label{tab_video_level_performance}
\resizebox{0.8\textwidth}{!}{
\begin{tabular}{cllllllllll}
\toprule
\multirow{3}{*}{Aspects} & \multirow{3}{*}{Methods} & \multicolumn{3}{c}{LGVQ} & \multicolumn{3}{c}{FETV} & \multicolumn{3}{c}{MQT} \\ 
\cmidrule(lr){3-5} \cmidrule(lr){6-8} \cmidrule(lr){9-11}
& & SRCC & KRCC & PLCC & SRCC & KRCC & PLCC & SRCC & KRCC & PLCC \\
\midrule
\multirow{6}{*}{Spatial} 
& UNIQUE~\cite{UNIQUE}           & 0.716             & 0.525             & 0.768             & 0.764             & 0.637             & 0.794             & - & - & - \\
& MUSIQ~\cite{MUSIQ}             & 0.669             & 0.491             & 0.682             & 0.722             & 0.613             & 0.758             & - & - & - \\
& StairIQA~\cite{StairIQA}       & 0.701             & 0.521             & 0.737             & \underline{0.806} & \underline{0.643} & \underline{0.812} & - & - & - \\
& CLIP-IQA~\cite{CLIP_IQA}       & 0.684             & 0.502             & 0.709             & 0.741             & 0.619             & 0.767             & - & - & - \\
& LIQE~\cite{LIQE}               & \underline{0.721} & \underline{0.538} & \underline{0.752} & 0.765             & 0.635             & 0.799             & - & - & - \\
& \textbf{Ours}                  & \textbf{0.764}    & \textbf{0.571}    & \textbf{0.793}    & \textbf{0.841}    & \textbf{0.685}    & \textbf{0.841}    & - & - & - \\
\midrule
\multirow{7}{*}{Temporal} 
& TLVQM~\cite{TLVQM}             & 0.828             & 0.616             & 0.832             & 0.825             & 0.675             & 0.837             & 0.813             & 0.605             & 0.831             \\
& RAPIQUE~\cite{RAPIQUE}         & 0.836             & 0.641             & 0.851             & 0.833             & 0.691             & 0.854             & 0.822             & 0.627             & 0.837             \\
& VSFA~\cite{VSFA}               & 0.841             & 0.643             & 0.857             & 0.839             & 0.705             & 0.859             & 0.834             & 0.630             & 0.851             \\
& SimpleVQA~\cite{simpleVQA}     & 0.857             & 0.659             & 0.867             & 0.852             & 0.726             & 0.862             & 0.848             & 0.644             & 0.856             \\
& FastVQA~\cite{FAST_VQA}        & 0.849             & 0.647             & 0.843             & 0.842             & 0.714             & 0.847             & 0.842             & 0.638             & 0.849             \\
& DOVER~\cite{dover}             & \underline{0.867} & \underline{0.672} & \underline{0.878} & \underline{0.868} & \underline{0.731} & \underline{0.881} & \underline{0.854} & \underline{0.665} & \underline{0.869} \\
& \textbf{Ours}                  & \textbf{0.894}    & \textbf{0.703}    & \textbf{0.910}    & \textbf{0.897}    & \textbf{0.753}    & \textbf{0.907}    & \textbf{0.898}    & \textbf{0.733}    & \textbf{0.909}    \\
\midrule
\multirow{8}{*}{Alignment} 
& CLIPScore~\cite{CLIPScore}     & 0.446             & 0.301             & 0.453             & 0.607             & 0.498             & 0.633             & 0.772             & 0.611             & 0.783             \\
& BLIP~\cite{BLIP}               & 0.455             & 0.319             & 0.464             & 0.616             & 0.505             & 0.645             & 0.761             & 0.616             & 0.772             \\
& viCLIP~\cite{viCLIP}           & 0.479             & 0.338             & 0.487             & 0.628             & 0.518             & 0.652             & 0.798             & 0.628             & 0.818             \\
& ImageReward~\cite{ImageReward} & 0.498             & 0.344             & 0.499             & 0.657             & 0.519             & 0.687             & 0.794             & 0.624             & 0.812             \\
& PickScore~\cite{PickScore}     & 0.501             & 0.353             & \underline{0.515} & 0.669             & 0.533             & \underline{0.708} & \underline{0.823} & \underline{0.649} & \underline{0.831} \\
& HPSv1~\cite{HPSv1}             & 0.481             & 0.341             & 0.497             & 0.639             & 0.525             & 0.680             & 0.781             & 0.620             & 0.785             \\
& HPSv2~\cite{HPSv2}             & \underline{0.504} & \underline{0.357} & 0.511             & \underline{0.686} & \underline{0.540} & 0.703             & 0.819             & 0.643             & 0.821             \\
& \textbf{Ours}                  & \textbf{0.545}    & \textbf{0.391}    & \textbf{0.569}    & \textbf{0.734}    & \textbf{0.572}    & \textbf{0.737}    & \textbf{0.845}    & \textbf{0.668}    & \textbf{0.851}    \\
\bottomrule
\end{tabular}
}
\end{table*}

\subsubsection{Evaluation Datasets}
We validate our proposed method on \textbf{LGVQ}, and two publicly available AIGC VQA datasets, \textbf{FETV}~\cite{FETV} and \textbf{MQT}~\cite{MQT}. The FETV dataset contains $2,476$ videos, which were generated by $619$ prompts and $4$ T2V models. Each video was subjectively rated by $3$ annotators across $4$ quality dimensions: static quality, temporal quality, overall alignment, and fine-grained alignment. Notably, overall alignment is heavily influenced by fine-grained alignment due to their high correlation. Therefore, in this study, we focus on three quality dimensions: static quality, temporal quality, and overall alignment for our experiments, aligning them with the corresponding quality dimensions in LGVQ. 
The MQT~\cite{MQT} dataset consists of $1,005$ videos generated from $201$ prompts and $5$ T2V models. Each video was subjectively rated by $24$ annotators on $2$ quality dimensions: alignment and perception. For our experiments, we adjust the number of neurons in the quality regression module to predict alignment and perception quality scores for the MQT~\cite{MQT}.


To ensure a fair comparison of the performance between the proposed UGVQ and other compared quality metrics, we adopt a train/validation/test split to retrain all metrics. The final model is selected based on its best performance on the validation set and subsequently evaluated on the test set. The reported results are averaged over $10$ trials to ensure a reliable measure of the method’s generalization capability. In our experiments, the data is split into training, validation, and test sets in a ratio of approximately $7:1:2$. It is important to note that each dataset contains multiple videos generated from the same prompt. To ensure our method generalizes well to unseen prompts, we adopt an \textbf{invisible prompt} strategy, where no prompt in the validation or test sets appears in the training set. That is, the prompts used for training are completely separated from those used for validation and testing.

\subsubsection{Compared Quality Metrics} 
To validate the performance of the proposed UGVQ metric, we select a range of popular learnable quality metrics as baseline models. These include:
\begin{itemize}
\item Five IQA metrics for spatial quality assessment: UNIQUE~\cite{UNIQUE}, MUSIQ~\cite{MUSIQ}, StairIQA~\cite{StairIQA}, CLIP-IQA~\cite{CLIP_IQA}, and LIQE~\cite{LIQE}.
\item Six VQA metrics for temporal quality assessment: TLVQM~\cite{TLVQM}, RAPIQUE~\cite{RAPIQUE}, VSFA~\cite{VSFA}, SimpleVQA~\cite{simpleVQA}, FastVQA~\cite{FAST_VQA}, and DOVER~\cite{dover}.
\item Seven CLIP-based methods for text-video alignment: CLIP~\cite{CLIP}, BLIP~\cite{BLIP}, viCLIP~\cite{viCLIP}, ImageReward~\cite{ImageReward}, PickScore~\cite{PickScore}, HPSv1~\cite{HPSv1}, and HPSv2~\cite{HPSv2}.
\end{itemize}

\subsubsection{Implementation Details}


For the spatial feature extractor, the number of keyframes $N_s$ is set to $8$. The Transformer encoder used in the spatial feature extractor consists of $8$ layers, with each layer having $2$ attention heads and a feedforward network dimension of $2048$. For the multi-head attention in the feature fusion module, the number of heads and the feedforward network dimension are set to $4$ and $1536$ respectively. For the quality regressor, the number of hidden neurons in the MLP is set to $9216$.



We initialize the ViT of the spatial feature extractor with the weights pretrained on the Pick-a-pic dataset~\cite{PickScore}, and the SlowFast~\cite{SlowFast} of the temporal feature extractor with weights pre-trained on the Kinetics-400 dataset. The models are optimized using the Adam optimizer with an initial learning rate of $1 \times 10^{-5}$. A step-wise learning rate decay strategy is applied, where the learning rate is reduced by $90\%$ every $5$ epochs. The models are trained for $50$ epochs with a batch size of $32$. 

To ensure reproducibility, we set the same random seeds for NumPy, CUDA, and PyTorch across all experiments. All experiments are performed on an Intel(R) Xeon(R) Gold 6354 CPU @3.00GHz and an Nvidia GeForce RTX 3090 GPU.
Similar to the experiments in Section~\ref{sec_benchmark}, the performance of the quality metrics is evaluated using three standard indices---SRCC, KRCC, and PLCC---at both video-level and model-level.

\subsection{Performance Comparison}

\begin{table*}[]
\caption{The model-level performance of the proposed UGVQ metric and the compared quality metrics on the LGVQ, FETV, and MQT datasets. The best-performing metric is highlighted in bold, while the second-best metric is underlined.}
\label{tab_model_level_performance}
\resizebox{0.8\textwidth}{!}{
\begin{tabular}{cllllllllll}
\toprule
\multirow{3}{*}{Aspects} & \multirow{3}{*}{Methods} & \multicolumn{3}{c}{LGVQ} & \multicolumn{3}{c}{FETV} & \multicolumn{3}{c}{MQT} \\ 
\cmidrule(lr){3-5} \cmidrule(lr){6-8} \cmidrule(lr){9-11}
& & SRCC & KRCC & PLCC & SRCC & KRCC & PLCC & SRCC & KRCC & PLCC \\
\midrule
\multirow{6}{*}{Spatial} 
& UNIQUE~\cite{UNIQUE}           & 0.937             & 0.905             & 0.965             & 0.899             & 0.812             & 0.915             & - & - & - \\
& MUSIQ~\cite{MUSIQ}             & 0.914             & 0.821             & 0.937             & 0.868             & 0.785             & 0.888             & - & - & - \\
& StairIQA~\cite{StairIQA}       & 0.928             & 0.867             & 0.941             & \underline{0.947} & \underline{0.919} & \underline{0.976} & - & - & - \\
& CLIP-IQA~\cite{CLIP_IQA}       & 0.917             & 0.835             & 0.946             & 0.877             & 0.855             & 0.896             & - & - & - \\
& LIQE~\cite{LIQE}               & \underline{0.955} & \underline{0.911} & \underline{0.967} & 0.925             & 0.899             & 0.949             & - & - & - \\
& \textbf{Ours}                  & \textbf{0.971}    & \textbf{0.933}    & \textbf{0.996}    & \textbf{0.972}    & \textbf{0.953}    & \textbf{0.989}    & - & - & - \\
\midrule
\multirow{7}{*}{Temporal} 
& TLVQM~\cite{TLVQM}             & 0.879             & 0.785             & 0.922             & 0.949             & 0.903             & 0.958             & 0.858             & 0.759             & 0.886            \\
& RAPIQUE~\cite{RAPIQUE}         & 0.924             & 0.835             & 0.966             & 0.944             & 0.908             & 0.966             & 0.871             & 0.732             & 0.884            \\
& VSFA~\cite{VSFA}               & 0.904             & 0.809             & 0.953             & 0.955             & 0.928             & 0.971             & 0.885             & 0.769             & 0.908            \\
& SimpleVQA~\cite{simpleVQA}     & 0.926             & 0.835             & 0.976             & 0.971             & 0.961             & 0.982             & 0.896             & 0.782             & 0.910            \\
& FastVQA~\cite{FAST_VQA}        & \underline{0.930} & \underline{0.861} & \underline{0.985} & 0.958             & 0.938             & 0.966             & 0.888             & 0.774             & 0.913            \\
& DOVER~\cite{dover}             & 0.918             & 0.823             & 0.966             & \underline{0.981} & \underline{0.962} & \underline{0.994} & \underline{0.903} & \underline{0.797} & \underline{0.921}\\
& \textbf{Ours}                  & \textbf{0.942}    & \textbf{0.866}    & \textbf{0.998}    & \textbf{0.987}    & \textbf{0.966}    & \textbf{0.999}    & \textbf{0.934}    & \textbf{0.802}    & \textbf{0.953}   \\
\midrule
\multirow{8}{*}{Alignment} 
& CLIPScore~\cite{CLIPScore}     & 0.894             & 0.822             & 0.907             & 0.827             & 0.740             & 0.886             & 0.833             & 0.698             & 0.866             \\
& BLIP~\cite{BLIP}               & 0.881             & 0.793             & 0.895             & 0.855             & 0.803             & 0.904             & 0.824             & 0.692             & 0.856             \\
& viCLIP~\cite{viCLIP}           & 0.916             & 0.888             & 0.942             & 0.847             & 0.775             & 0.899             & 0.862             & 0.706             & 0.890             \\
& ImageReward~\cite{ImageReward} & 0.934             & 0.914             & 0.955             & 0.871             & 0.854             & 0.929             & 0.856             & 0.713             & 0.886             \\
& PickScore~\cite{PickScore}     & \underline{0.947} & \underline{0.920} & \underline{0.976} & 0.917             & 0.872             & 0.974             & \underline{0.890} & \underline{0.725} & \underline{0.910} \\
& HPSv1~\cite{HPSv1}             & 0.888             & 0.807             & 0.902             & 0.852             & 0.806             & 0.901             & 0.851             & 0.716             & 0.876             \\
& HPSv2~\cite{HPSv2}             & 0.906             & 0.872             & 0.932             & \underline{0.920} & \underline{0.876} & \underline{0.975} & 0.876             & 0.709             & 0.901             \\
& \textbf{Ours}                  & \textbf{0.977}    & \textbf{0.947}    & \textbf{0.989}    & \textbf{0.945}    & \textbf{0.909}    & \textbf{0.991}    & \textbf{0.900}    & \textbf{0.748}    & \textbf{0.933}    \\
\bottomrule
\end{tabular}
}
\end{table*}


The experimental results at the video-level and model-level are presented in Table~\ref{tab_video_level_performance} and Table~\ref{tab_model_level_performance}, respectively. \textbf{First}, we observe that our proposed UGVQ metric achieves the best performance across all three quality dimensions on all tested datasets compared to other quality metrics, which verifies the effectiveness of the model design of UGVQ. \textbf{Second}, UGVQ demonstrates significantly better performance in assessing temporal quality compared to spatial quality, with both outperforming text-video alignment by a large margin. This suggests an increasing level of difficulty in quality assessment among the three dimensions: temporal quality, spatial quality, and text-video alignment. This trend may be attributed to the relative complexity of each dimension—--temporal quality assessment involves evaluating feature variations across frames, spatial quality requires analyzing complex AIGC artifacts and content, and text-video alignment further necessitates understanding the semantic content of both text prompts and generated videos. \textbf{Third}, UGVQ achieves consistently high performance across all three quality dimensions in the model-level evaluation, with SRCC values exceeding $0.9$. This indicates a strong alignment with human perception and highlights UGVQ's capability in assessing the generation performance of T2V models.

\subsection{Ablation Study}
\label{experiment_ablation}

\begin{table*}
  \caption{Ablation study of the proposed UGVQ methods. The spatial feature extractor, temporal feature extractor, text feature extractor, and feature fusion module are abbreviated as Spatial, Temporal, Text, and Fusion respectively.}
  \label{tab_ablation}
  \resizebox{0.9\textwidth}{!}{
  \begin{tabular}{ccccc|ccc ccc ccc}
    \toprule
                             & \multicolumn{4}{c}{Features}                  & \multicolumn{3}{c}{Spatial}       & \multicolumn{3}{c}{Temporal}      & \multicolumn{3}{c}{Alignment} \\
                             \cmidrule(r){2-5}                               \cmidrule(r){6-8}                   \cmidrule(r){9-11}                  \cmidrule(r){12-14}               
    \multicolumn{1}{c}{No.}  & Spatial   & Temporal  & Text      & Fusion    & SRCC      & KRCC      & PLCC      & SRCC      & KRCC      & PLCC      & SRCC      & KRCC      & PLCC  \\
    \multicolumn{1}{c}{1}    & \ding{52} &           &           &           & 0.712     & 0.568     & 0.745     & 0.821     & 0.632     & 0.837     & 0.483     & 0.358     & 0.526 \\
    \multicolumn{1}{c}{2}    &           & \ding{52} &           &           & 0.581     & 0.433     & 0.628     & 0.876     & 0.683     & 0.892     & 0.471     & 0.351     & 0.496 \\
    \multicolumn{1}{c}{3}    &           &           & \ding{52} &           & 0.248     & 0.222     & 0.315     & 0.222     & 0.151     & 0.238     & 0.308     & 0.213     & 0.334 \\
    \multicolumn{1}{c}{4}    &           & \ding{52} & \ding{52} &           & 0.687     & 0.559     & 0.781     & 0.889     & 0.697     & 0.902     & 0.525     & 0.376     & 0.548 \\
    \multicolumn{1}{c}{5}    & \ding{52} &           & \ding{52} &           & 0.752     & 0.563     & 0.783     & 0.854     & 0.668     & 0.869     & 0.508     & 0.365     & 0.530 \\
    \multicolumn{1}{c}{6}    & \ding{52} & \ding{52} &           &           & 0.745     & 0.558     & 0.779     & 0.886     & 0.692     & 0.895     & 0.516     & 0.372     & 0.537 \\
    \multicolumn{1}{c}{7}    & \ding{52} & \ding{52} & \ding{52} &           & 0.753     & 0.563     & 0.785     & 0.886     & 0.694     & 0.903     & 0.530     & 0.382     & 0.558 \\
    \multicolumn{1}{c}{8}    & \ding{52} & \ding{52} & \ding{52} & \ding{52} & 0.764     & 0.571     & 0.793     & 0.894     & 0.703     & 0.910     & 0.545     & 0.391     & 0.569 \\
    \bottomrule
  \end{tabular}
  }
  \centering
\end{table*}

In this section, we conduct an ablation study to evaluate the effectiveness of the four components of our UGVQ method: the spatial feature extractor, the temporal feature extractor, the text feature extractor, and the feature fusion module. The experimental results are presented in Table ~\ref{tab_ablation}, where each combination of components is evaluated using SRCC, KRCC, and PLCC indices across the spatial quality, temporal quality, and text-video alignment dimensions.

\textbf{First}, when comparing the performance of the three individual feature extraction modules, we observe that the spatial feature extractor performs the best on the spatial quality assessment, while the temporal feature extractor performs the best on the temporal quality assessment. This demonstrates that the proposed feature extraction modules are well-suited for evaluating their respective quality dimensions. As for the text feature extractor, since it only extracts features from text prompts, it cannot independently evaluate any of the three quality dimensions. However, when text features are combined with spatial or temporal features, they significantly enhance text-video alignment performance. \textbf{Second}, when evaluating the performance of combined feature extraction modules, we observe a consistent performance improvement compared to the individual feature extraction modules. For example, the combination of the spatial and temporal feature extractors outperforms the individual spatial and temporal feature extractors in terms of spatial quality, temporal quality, and text-video alignment. This demonstrates that the proposed features are complementary and collectively contribute to improving AIGC video quality assessment. \textbf{Third}, when combining all three types of features, the model achieves the best performance compared to other combinations. Furthermore, incorporating the proposed feature fusion module further enhances the overall model performance. These results demonstrate the rationality and effectiveness of the UGVQ model design.

\subsection{Cross-Dataset Evaluation}

\begin{table}
\caption{The performance of the proposed UGVQ metirc and the compared quality metrics on cross-dataset evaluation. The best-performing metric is highlighted in bold.}
\label{tab_cross_dataset}
\resizebox{0.9\textwidth}{!}{
    \begin{tabular}{clllllllllllll}
    \toprule
    \multirow{3}{*}{Aspects}   & \multicolumn{1}{l}{\multirow{3}{*}{Methods}}  & \multicolumn{6}{c}{$LGVQ \rightarrow FETV$}                                                             & \multicolumn{6}{c}{$FETV \rightarrow LGVQ$}                                                         \\
                                                                               \cmidrule(lr){3-8}                                                                                        \cmidrule(lr){9-14}                                                                                     
                               &                                               & \multicolumn{3}{c}{Video-level}                    & \multicolumn{3}{c}{Model-level}                    & \multicolumn{3}{c}{Video-level}                   & \multicolumn{3}{c}{Model-level}                  \\
                                                                               \cmidrule(lr){3-5}                                   \cmidrule(lr){6-8}                                   \cmidrule(lr){9-11}                                  \cmidrule(lr){12-14}                                
                               & \multicolumn{1}{c}{}                          & SRCC           & KRCC           & PLCC             & SRCC           & KRCC           & PLCC             & SRCC           & KRCC           & PLCC            & SRCC           & KRCC           & PLCC            \\
    \midrule                                                                                                                                                                                                                                                                                    
    \multirow{6}{*}{Spatial}   & UNIQUE~\cite{UNIQUE}                          & 0.389          & 0.272          & 0.383            & 0.846          & 0.786          & 0.882            & 0.360          & 0.252          & 0.353           & 0.696          & 0.611          & 0.353            \\
                               & MUSIQ~\cite{MUSIQ}                            & 0.426          & 0.312          & 0.472            & 0.869          & 0.813          & 0.911            & 0.406          & 0.281          & 0.404           & 0.749          & 0.659          & 0.404            \\
                               & StairIQA~\cite{StairIQA}                      & 0.501          & 0.360          & 0.539            & 0.951          & 0.907          & 0.993            & 0.484          & 0.346          & 0.500           & 0.744          & 0.648          & 0.500            \\
                               & CLIP-IQA~\cite{CLIP_IQA}                      & 0.503          & 0.361          & 0.543            & 0.968          & 0.932          & 0.997            & 0.493          & 0.355          & 0.501           & 0.823          & 0.735          & 0.501            \\
                               & LIQE~\cite{LIQE}                              & 0.478          & 0.341          & 0.519            & 0.944          & 0.919          & 0.988            & 0.461          & 0.340          & 0.477           & 0.735          & 0.645          & 0.477            \\
                               & \textbf{Ours}                                 & \textbf{0.553} & \textbf{0.406} & \textbf{0.555}   & \textbf{0.998} & \textbf{0.966} & \textbf{0.999}   & \textbf{0.521} & \textbf{0.359} & \textbf{0.524}  & \textbf{0.828} & \textbf{0.733} & \textbf{0.962}   \\
    \midrule                                                                                                                                                                                                                                                                                    
    \multirow{7}{*}{Temporal}  & TLVQM~\cite{TLVQM}                            & 0.306          & 0.211          & 0.314            & 0.825          & 0.745          & 0.872            & 0.310          & 0.211          & 0.314           & 0.622          & 0.544          & 0.314            \\
                               & RAPIQUE~\cite{RAPIQUE}                        & 0.373          & 0.260          & 0.351            & 0.857          & 0.783          & 0.898            & 0.347          & 0.247          & 0.351           & 0.643          & 0.571          & 0.351            \\
                               & VSFA~\cite{VSFA}                              & 0.396          & 0.272          & 0.364            & 0.911          & 0.878          & 0.936            & 0.388          & 0.274          & 0.398           & 0.691          & 0.609          & 0.398            \\
                               & SimpleVQA~\cite{simpleVQA}                    & 0.501          & 0.366          & 0.511            & 0.930          & 0.904          & 0.959            & 0.419          & 0.279          & 0.407           & 0.677          & 0.589          & 0.407            \\
                               & FastVQA~\cite{FAST_VQA}                       & 0.482          & 0.324          & 0.494            & 0.957          & 0.922          & 0.983            & 0.397          & 0.272          & 0.364           & 0.682          & 0.609          & 0.364            \\
                               & DOVER~\cite{dover}                            & 0.494          & 0.349          & 0.483            & 0.973          & 0.947          & 0.996            & 0.427          & 0.287          & 0.406           & 0.736          & 0.657          & 0.406            \\
                               & \textbf{Ours}                                 & \textbf{0.512} & \textbf{0.368} & \textbf{0.535}   & \textbf{0.999} & \textbf{0.967} & \textbf{0.999}   & \textbf{0.442} & \textbf{0.296} & \textbf{0.432}  & \textbf{0.748} & \textbf{0.666} & \textbf{0.894}   \\
    \midrule                                                                                                                                                                                                                                                                                    
    \multirow{8}{*}{Alignment} & CLIPScore~\cite{CLIPScore}                    & 0.234          & 0.167          & 0.252            & 0.334          & 0.267          & 0.622            & 0.168          & 0.112          & 0.205           & 0.326          & 0.285          & 0.731           \\
                               & BLIP~\cite{BLIP}                              & 0.216          & 0.144          & 0.233            & 0.331          & 0.278          & 0.619            & 0.151          & 0.103          & 0.193           & 0.304          & 0.263          & 0.726           \\
                               & viCLIP~\cite{viCLIP}                          & 0.253          & 0.173          & 0.274            & 0.384          & 0.325          & 0.655            & 0.177          & 0.117          & 0.212           & 0.338          & 0.301          & 0.711            \\
                               & ImageReward~\cite{ImageReward}                & 0.261          & 0.183          & 0.283            & 0.389          & 0.327          & 0.701            & 0.193          & 0.135          & 0.245           & 0.353          & 0.315          & 0.739           \\
                               & PickScore~\cite{PickScore}                    & 0.259          & 0.178          & 0.284            & 0.397          & 0.333          & 0.725            & 0.181          & 0.124          & 0.229           & 0.357          & 0.309          & 0.757            \\
                               & HPSv1~\cite{HPSv1}                            & 0.228          & 0.152          & 0.248            & 0.340          & 0.281          & 0.663            & 0.153          & 0.104          & 0.195           & 0.303          & 0.267          & 0.698           \\
                               & HPSv2~\cite{HPSv2}                            & 0.263          & 0.189          & 0.285            & 0.384          & 0.323          & 0.716            & 0.201          & 0.140          & 0.243           & 0.342          & 0.298          & 0.727           \\
                               & \textbf{Ours}                                 & \textbf{0.278} & \textbf{0.196} & \textbf{0.292}   & \textbf{0.399} & \textbf{0.333} & \textbf{0.730}   & \textbf{0.217} & \textbf{0.148} & \textbf{0.255}  & \textbf{0.371} & \textbf{0.333} & \textbf{0.769}  \\
    \bottomrule
    \end{tabular}
}
\centering
\end{table}


To further assess the generalization capability of UGVQ, we conduct cross-dataset experiments under two settings: \textbf{(1) training on the LGVQ dataset and evaluating on the FETV dataset}, and \textbf{(2) training on the FETV dataset and evaluating on the LGVQ dataset}. Table~\ref{tab_cross_dataset} presents the results for both video-level and model-level evaluations. For fair comparison, we also report the performance of competing quality assessment metrics.

From Table~\ref{tab_cross_dataset}, we observe three key findings. \textbf{First}, our UGVQ metric consistently outperforms competing metrics across all three quality dimensions in both cross-dataset settings. This demonstrates that UGVQ effectively learns better quality feature representations for AIGC video quality assessment. \textbf{Second}, the performance of the $LGVQ \rightarrow FETV$ setting is consistently better than that of the $FETV \rightarrow LGVQ$ setting for nearly all quality metrics. This can be attributed to the higher precision of LGVQ quality labels compared to FETV. Specifically, each video in FETV was rated by only $3$ subjects, whereas in LGVQ, each video was rated by $10$ subjects. As a result, LGVQ provides richer supervision, enabling the model to learn better feature representations for AIGC videos. This also indicates that LGVQ is more suitable for training AIGC video evaluators. \textbf{Third}, in the $LGVQ \rightarrow FETV$ setting, we find that the SRCC value for model-level evaluation exceeds $0.99$ for both spatial and temporal quality dimensions. This further demonstrates that our UGVQ metric can effectively evaluate the generation performance of T2V model. However, for text-video alignment, we observe consistently low correlations across all quality metrics, regardless of whether the evaluation is conducted at the video-level or model-level. This suggests that text-video alignment remains a challenging task in AIGC video evaluation.

\section{Conclusion}


In this paper, we introduce LGVQ, a multi-dimensional quality assessment dataset for AIGC videos, comprising $2,808$ AIGC videos generated by six T2V generation methods using $468$ text prompts. We conduct a subjective quality assessment experiment on LGVQ, evaluating the videos across three critical dimensions: spatial quality, temporal quality, and text-video alignment. Additionally, we benchmark existing quality assessment metrics on the LGVQ dataset, revealing their limitations in measuring the perceptual quality of AIGC videos. To address these challenges, we propose UGVQ metric, designed to simultaneously evaluate all three quality dimensions. Extensive experimental results demonstrate that UGVQ significantly outperforms existing quality metrics, verifying its effectiveness as a robust and comprehensive evaluation tool for AIGC videos.

However, it should be noted that text-to-video generation is a rapidly evolving field, with new T2V models being released frequently. As a result, many outstanding T2V models that emerged after we conducted this study are not included in the LGVQ dataset. What's more, many commercial T2V tools such as Sora have also become publicly available. In future work, we plan to expand the LGVQ dataset by incorporating both state-of-the-art open-source T2V models and commercial T2V tools. This will enable a fairer comparison of the generation performance of different T2V models and allow for a more comprehensive analysis of their strengths and limitations. Additionally, our study highlights that text-video alignment remains a significant challenge in AIGC video evaluation. Moving forward, we will place greater emphasis on improving text-video alignment evaluation and enhancing the overall performance.

\bibliographystyle{ACM-Reference-Format}
\bibliography{sample-base}


\begin{thebibliography}{99}


\ifx \showCODEN    \undefined \def \showCODEN     #1{\unskip}     \fi
\ifx \showDOI      \undefined \def \showDOI       #1{#1}\fi
\ifx \showISBNx    \undefined \def \showISBNx     #1{\unskip}     \fi
\ifx \showISBNxiii \undefined \def \showISBNxiii  #1{\unskip}     \fi
\ifx \showISSN     \undefined \def \showISSN      #1{\unskip}     \fi
\ifx \showLCCN     \undefined \def \showLCCN      #1{\unskip}     \fi
\ifx \shownote     \undefined \def \shownote      #1{#1}          \fi
\ifx \showarticletitle \undefined \def \showarticletitle #1{#1}   \fi
\ifx \showURL      \undefined \def \showURL       {\relax}        \fi
\providecommand\bibfield[2]{#2}
\providecommand\bibinfo[2]{#2}
\providecommand\natexlab[1]{#1}
\providecommand\showeprint[2][]{arXiv:#2}

\bibitem[Met(2002)]%
        {Methodology}
 \bibinfo{year}{2002}\natexlab{}.
\newblock \showarticletitle{Methodology for the subjective assessment of the quality of television pictures}.
\newblock \bibinfo{journal}{\emph{International Telecommunication Union}} (\bibinfo{year}{2002}).
\newblock


\bibitem[Betker et~al\mbox{.}(2023)]%
        {betker2023improving}
\bibfield{author}{\bibinfo{person}{James Betker}, \bibinfo{person}{Gabriel Goh}, \bibinfo{person}{Li Jing}, \bibinfo{person}{Tim Brooks}, \bibinfo{person}{Jianfeng Wang}, \bibinfo{person}{Linjie Li}, \bibinfo{person}{Long Ouyang}, \bibinfo{person}{Juntang Zhuang}, \bibinfo{person}{Joyce Lee}, \bibinfo{person}{Yufei Guo}, {et~al\mbox{.}}} \bibinfo{year}{2023}\natexlab{}.
\newblock \showarticletitle{Improving image generation with better captions}.
\newblock \bibinfo{journal}{\emph{Computer Science}} \bibinfo{volume}{2}, \bibinfo{number}{3} (\bibinfo{year}{2023}), \bibinfo{pages}{8}.
\newblock


\bibitem[Binkowski et~al\mbox{.}(2018)]%
        {KVD}
\bibfield{author}{\bibinfo{person}{Mikolaj Binkowski}, \bibinfo{person}{Danica~J Sutherland}, \bibinfo{person}{Michael Arbel}, {and} \bibinfo{person}{Arthur Gretton}.} \bibinfo{year}{2018}\natexlab{}.
\newblock \showarticletitle{Towards Accurate Generative Models of Video: A New Metric \& Challenges}.
\newblock \bibinfo{journal}{\emph{arXiv preprint arXiv:1812.01717}} (\bibinfo{year}{2018}).
\newblock


\bibitem[Black et~al\mbox{.}(2023)]%
        {black2023training}
\bibfield{author}{\bibinfo{person}{Kevin Black}, \bibinfo{person}{Michael Janner}, \bibinfo{person}{Yilun Du}, \bibinfo{person}{Ilya Kostrikov}, {and} \bibinfo{person}{Sergey Levine}.} \bibinfo{year}{2023}\natexlab{}.
\newblock \showarticletitle{Training Diffusion Models with Reinforcement Learning}. In \bibinfo{booktitle}{\emph{The Twelfth International Conference on Learning Representations}}.
\newblock


\bibitem[Carreira and Zisserman(2017)]%
        {kinetics}
\bibfield{author}{\bibinfo{person}{Joao Carreira} {and} \bibinfo{person}{Andrew Zisserman}.} \bibinfo{year}{2017}\natexlab{}.
\newblock \showarticletitle{Quo vadis, action recognition? a new model and the kinetics dataset}. In \bibinfo{booktitle}{\emph{proceedings of the IEEE Conference on Computer Vision and Pattern Recognition}}. \bibinfo{pages}{6299--6308}.
\newblock


\bibitem[Chen et~al\mbox{.}(2024a)]%
        {TOPIQ}
\bibfield{author}{\bibinfo{person}{Chaofeng Chen}, \bibinfo{person}{Jiadi Mo}, \bibinfo{person}{Jingwen Hou}, \bibinfo{person}{Haoning Wu}, \bibinfo{person}{Liang Liao}, \bibinfo{person}{Wenxiu Sun}, \bibinfo{person}{Qiong Yan}, {and} \bibinfo{person}{Weisi Lin}.} \bibinfo{year}{2024}\natexlab{a}.
\newblock \showarticletitle{TOPIQ: A Top-Down Approach From Semantics to Distortions for Image Quality Assessment}.
\newblock \bibinfo{journal}{\emph{IEEE Transactions on Image Processing}}  \bibinfo{volume}{33} (\bibinfo{year}{2024}), \bibinfo{pages}{2404--2418}.
\newblock
\urldef\tempurl%
\url{https://doi.org/10.1109/TIP.2024.3378466}
\showDOI{\tempurl}


\bibitem[Chen et~al\mbox{.}(2023a)]%
        {chen2023videocrafter1}
\bibfield{author}{\bibinfo{person}{Haoxin Chen}, \bibinfo{person}{Menghan Xia}, \bibinfo{person}{Yingqing He}, \bibinfo{person}{Yong Zhang}, \bibinfo{person}{Xiaodong Cun}, \bibinfo{person}{Shaoshu Yang}, \bibinfo{person}{Jinbo Xing}, \bibinfo{person}{Yaofang Liu}, \bibinfo{person}{Qifeng Chen}, \bibinfo{person}{Xintao Wang}, {et~al\mbox{.}}} \bibinfo{year}{2023}\natexlab{a}.
\newblock \showarticletitle{Videocrafter1: Open diffusion models for high-quality video generation}.
\newblock \bibinfo{journal}{\emph{arXiv preprint arXiv:2310.19512}} (\bibinfo{year}{2023}).
\newblock


\bibitem[Chen et~al\mbox{.}(2023b)]%
        {VideoCrafter}
\bibfield{author}{\bibinfo{person}{Haoxin Chen}, \bibinfo{person}{Menghan Xia}, \bibinfo{person}{Yingqing He}, \bibinfo{person}{Yong Zhang}, \bibinfo{person}{Xiaodong Cun}, \bibinfo{person}{Shaoshu Yang}, \bibinfo{person}{Jinbo Xing}, \bibinfo{person}{Yaofang Liu}, \bibinfo{person}{Qifeng Chen}, \bibinfo{person}{Xintao Wang}, \bibinfo{person}{Chao Weng}, {and} \bibinfo{person}{Ying Shan}.} \bibinfo{year}{2023}\natexlab{b}.
\newblock \bibinfo{title}{VideoCrafter1: Open Diffusion Models for High-Quality Video Generation}.
\newblock
\newblock
\showeprint[arxiv]{2310.19512}~[cs.CV]


\bibitem[Chen et~al\mbox{.}(2024b)]%
        {GAIA}
\bibfield{author}{\bibinfo{person}{Zijian Chen}, \bibinfo{person}{Wei Sun}, \bibinfo{person}{Yuan Tian}, \bibinfo{person}{Jun Jia}, \bibinfo{person}{Zicheng Zhang}, \bibinfo{person}{Jiarui Wang}, \bibinfo{person}{Ru Huang}, \bibinfo{person}{Xiongkuo Min}, \bibinfo{person}{Guangtao Zhai}, {and} \bibinfo{person}{Wenjun Zhang}.} \bibinfo{year}{2024}\natexlab{b}.
\newblock \showarticletitle{GAIA: Rethinking Action Quality Assessment for AI-Generated Videos}.
\newblock \bibinfo{journal}{\emph{arXiv preprint arXiv:2406.06087}} (\bibinfo{year}{2024}).
\newblock


\bibitem[Chivileva et~al\mbox{.}(2023)]%
        {MQT}
\bibfield{author}{\bibinfo{person}{Iya Chivileva}, \bibinfo{person}{Philip Lynch}, \bibinfo{person}{Tomas~E Ward}, {and} \bibinfo{person}{Alan~F Smeaton}.} \bibinfo{year}{2023}\natexlab{}.
\newblock \showarticletitle{Measuring the Quality of Text-to-Video Model Outputs: Metrics and Dataset}.
\newblock \bibinfo{journal}{\emph{arXiv preprint arXiv:2309.08009}} (\bibinfo{year}{2023}).
\newblock


\bibitem[Cho et~al\mbox{.}(2024)]%
        {cho2024sora}
\bibfield{author}{\bibinfo{person}{Joseph Cho}, \bibinfo{person}{Fachrina~Dewi Puspitasari}, \bibinfo{person}{Sheng Zheng}, \bibinfo{person}{Jingyao Zheng}, \bibinfo{person}{Lik-Hang Lee}, \bibinfo{person}{Tae-Ho Kim}, \bibinfo{person}{Choong~Seon Hong}, {and} \bibinfo{person}{Chaoning Zhang}.} \bibinfo{year}{2024}\natexlab{}.
\newblock \showarticletitle{Sora as an agi world model? a complete survey on text-to-video generation}.
\newblock \bibinfo{journal}{\emph{arXiv preprint arXiv:2403.05131}} (\bibinfo{year}{2024}).
\newblock


\bibitem[Deng et~al\mbox{.}(2019)]%
        {deng2019irc}
\bibfield{author}{\bibinfo{person}{Kangle Deng}, \bibinfo{person}{Tianyi Fei}, \bibinfo{person}{Xin Huang}, {and} \bibinfo{person}{Yuxin Peng}.} \bibinfo{year}{2019}\natexlab{}.
\newblock \showarticletitle{IRC-GAN: Introspective Recurrent Convolutional GAN for Text-to-video Generation.}. In \bibinfo{booktitle}{\emph{IJCAI}}. \bibinfo{pages}{2216--2222}.
\newblock


\bibitem[Ding et~al\mbox{.}(2022)]%
        {cogview2}
\bibfield{author}{\bibinfo{person}{Ming Ding}, \bibinfo{person}{Wendi Zheng}, \bibinfo{person}{Wenyi Hong}, {and} \bibinfo{person}{Jie Tang}.} \bibinfo{year}{2022}\natexlab{}.
\newblock \showarticletitle{CogView2: Faster and Better Text-to-Image Generation via Hierarchical Transformers}.
\newblock \bibinfo{journal}{\emph{arXiv preprint arXiv:2204.14217}} (\bibinfo{year}{2022}).
\newblock


\bibitem[Dosovitskiy et~al\mbox{.}(2021)]%
        {vit}
\bibfield{author}{\bibinfo{person}{Alexey Dosovitskiy}, \bibinfo{person}{Lucas Beyer}, \bibinfo{person}{Alexander Kolesnikov}, \bibinfo{person}{Dirk Weissenborn}, \bibinfo{person}{Xiaohua Zhai}, \bibinfo{person}{Thomas Unterthiner}, \bibinfo{person}{Mostafa Dehghani}, \bibinfo{person}{Matthias Minderer}, \bibinfo{person}{Georg Heigold}, \bibinfo{person}{Sylvain Gelly}, \bibinfo{person}{Jakob Uszkoreit}, {and} \bibinfo{person}{Neil Houlsby}.} \bibinfo{year}{2021}\natexlab{}.
\newblock \bibinfo{title}{An Image is Worth 16x16 Words: Transformers for Image Recognition at Scale}.
\newblock
\newblock
\showeprint[arxiv]{2010.11929}~[cs.CV]
\urldef\tempurl%
\url{https://arxiv.org/abs/2010.11929}
\showURL{%
\tempurl}


\bibitem[Du et~al\mbox{.}(2023)]%
        {aigc_ad}
\bibfield{author}{\bibinfo{person}{Duo Du}, \bibinfo{person}{Yanling Zhang}, {and} \bibinfo{person}{Jiao Ge}.} \bibinfo{year}{2023}\natexlab{}.
\newblock \showarticletitle{Effect of AI Generated Content Advertising on Consumer Engagement}. In \bibinfo{booktitle}{\emph{International Conference on Human-Computer Interaction}}. Springer, \bibinfo{pages}{121--129}.
\newblock


\bibitem[Esser et~al\mbox{.}(2023a)]%
        {esser2023structure}
\bibfield{author}{\bibinfo{person}{Patrick Esser}, \bibinfo{person}{Johnathan Chiu}, \bibinfo{person}{Parmida Atighehchian}, \bibinfo{person}{Jonathan Granskog}, {and} \bibinfo{person}{Anastasis Germanidis}.} \bibinfo{year}{2023}\natexlab{a}.
\newblock \showarticletitle{Structure and content-guided video synthesis with diffusion models}. In \bibinfo{booktitle}{\emph{Proceedings of the IEEE/CVF International Conference on Computer Vision}}. \bibinfo{pages}{7346--7356}.
\newblock


\bibitem[Esser et~al\mbox{.}(2023b)]%
        {Gen2}
\bibfield{author}{\bibinfo{person}{Patrick Esser}, \bibinfo{person}{Johnathan Chiu}, \bibinfo{person}{Parmida Atighehchian}, \bibinfo{person}{Jonathan Granskog}, {and} \bibinfo{person}{Anastasis Germanidis}.} \bibinfo{year}{2023}\natexlab{b}.
\newblock \showarticletitle{Structure and Content-Guided Video Synthesis with Diffusion Models}. In \bibinfo{booktitle}{\emph{Proceedings of the IEEE/CVF International Conference on Computer Vision (ICCV)}}. \bibinfo{pages}{7346--7356}.
\newblock


\bibitem[Fang et~al\mbox{.}(2023)]%
        {fang2023study}
\bibfield{author}{\bibinfo{person}{Yuming Fang}, \bibinfo{person}{Zhaoqian Li}, \bibinfo{person}{Jiebin Yan}, \bibinfo{person}{Xiangjie Sui}, {and} \bibinfo{person}{Hantao Liu}.} \bibinfo{year}{2023}\natexlab{}.
\newblock \showarticletitle{Study of spatio-temporal modeling in video quality assessment}.
\newblock \bibinfo{journal}{\emph{IEEE Transactions on Image Processing}} (\bibinfo{year}{2023}).
\newblock


\bibitem[Fang et~al\mbox{.}(2020)]%
        {SPAQ}
\bibfield{author}{\bibinfo{person}{Yuming Fang}, \bibinfo{person}{Hanwei Zhu}, \bibinfo{person}{Yan Zeng}, \bibinfo{person}{Kede Ma}, {and} \bibinfo{person}{Zhou Wang}.} \bibinfo{year}{2020}\natexlab{}.
\newblock \showarticletitle{Perceptual quality assessment of smartphone photography}. In \bibinfo{booktitle}{\emph{Proceedings of the IEEE/CVF conference on computer vision and pattern recognition}}. \bibinfo{pages}{3677--3686}.
\newblock


\bibitem[Feichtenhofer et~al\mbox{.}(2019)]%
        {SlowFast}
\bibfield{author}{\bibinfo{person}{Christoph Feichtenhofer}, \bibinfo{person}{Haoqi Fan}, \bibinfo{person}{Jitendra Malik}, {and} \bibinfo{person}{Kaiming He}.} \bibinfo{year}{2019}\natexlab{}.
\newblock \showarticletitle{SlowFast Networks for Video Recognition}. In \bibinfo{booktitle}{\emph{2019 IEEE/CVF International Conference on Computer Vision (ICCV)}}. \bibinfo{pages}{6201--6210}.
\newblock
\urldef\tempurl%
\url{https://doi.org/10.1109/ICCV.2019.00630}
\showDOI{\tempurl}


\bibitem[Ge et~al\mbox{.}(2024)]%
        {LMM_VQA}
\bibfield{author}{\bibinfo{person}{Qihang Ge}, \bibinfo{person}{Wei Sun}, \bibinfo{person}{Yu Zhang}, \bibinfo{person}{Yunhao Li}, \bibinfo{person}{Zhongpeng Ji}, \bibinfo{person}{Fengyu Sun}, \bibinfo{person}{Shangling Jui}, \bibinfo{person}{Xiongkuo Min}, {and} \bibinfo{person}{Guangtao Zhai}.} \bibinfo{year}{2024}\natexlab{}.
\newblock \showarticletitle{LMM-VQA: Advancing Video Quality Assessment with Large Multimodal Models}.
\newblock \bibinfo{journal}{\emph{arXiv preprint arXiv:2408.14008}} (\bibinfo{year}{2024}).
\newblock


\bibitem[Ghosh et~al\mbox{.}(2024)]%
        {Geneval}
\bibfield{author}{\bibinfo{person}{Dhruba Ghosh}, \bibinfo{person}{Hannaneh Hajishirzi}, {and} \bibinfo{person}{Ludwig Schmidt}.} \bibinfo{year}{2024}\natexlab{}.
\newblock \showarticletitle{Geneval: An object-focused framework for evaluating text-to-image alignment}.
\newblock \bibinfo{journal}{\emph{Advances in Neural Information Processing Systems}}  \bibinfo{volume}{36} (\bibinfo{year}{2024}).
\newblock


\bibitem[Golestaneh et~al\mbox{.}(2022)]%
        {TReS}
\bibfield{author}{\bibinfo{person}{S.~Alireza Golestaneh}, \bibinfo{person}{Saba Dadsetan}, {and} \bibinfo{person}{Kris~M. Kitani}.} \bibinfo{year}{2022}\natexlab{}.
\newblock \showarticletitle{No-Reference Image Quality Assessment via Transformers, Relative Ranking, and Self-Consistency}. In \bibinfo{booktitle}{\emph{Proceedings of the IEEE/CVF Winter Conference on Applications of Computer Vision (WACV)}}. \bibinfo{pages}{1220--1230}.
\newblock


\bibitem[Goyal et~al\mbox{.}(2017)]%
        {VQAv2}
\bibfield{author}{\bibinfo{person}{Yash Goyal}, \bibinfo{person}{Ammar Khattak}, \bibinfo{person}{Sandeep Kottur}, \bibinfo{person}{Amit Agrawal}, \bibinfo{person}{Dhruv Batra}, {and} \bibinfo{person}{Devi Parikh}.} \bibinfo{year}{2017}\natexlab{}.
\newblock \showarticletitle{Making the vqa model smarter: Learning from the web}. In \bibinfo{booktitle}{\emph{Proceedings of the IEEE International Conference on Computer Vision (ICCV)}}. \bibinfo{pages}{1759--1767}.
\newblock


\bibitem[Gu et~al\mbox{.}(2018)]%
        {Ava}
\bibfield{author}{\bibinfo{person}{Chunhui Gu}, \bibinfo{person}{Chen Sun}, \bibinfo{person}{David~A Ross}, \bibinfo{person}{Carl Vondrick}, \bibinfo{person}{Caroline Pantofaru}, \bibinfo{person}{Yeqing Li}, \bibinfo{person}{Sudheendra Vijayanarasimhan}, \bibinfo{person}{George Toderici}, \bibinfo{person}{Susanna Ricco}, \bibinfo{person}{Rahul Sukthankar}, {et~al\mbox{.}}} \bibinfo{year}{2018}\natexlab{}.
\newblock \showarticletitle{Ava: A video dataset of spatio-temporally localized atomic visual actions}. In \bibinfo{booktitle}{\emph{Proceedings of the IEEE conference on computer vision and pattern recognition}}. \bibinfo{pages}{6047--6056}.
\newblock


\bibitem[Gu et~al\mbox{.}(2023)]%
        {aigc_film}
\bibfield{author}{\bibinfo{person}{Rongzhang Gu}, \bibinfo{person}{Hui Li}, \bibinfo{person}{Changyue Su}, {and} \bibinfo{person}{Wenyan Wu}.} \bibinfo{year}{2023}\natexlab{}.
\newblock \showarticletitle{Innovative Digital Storytelling with AIGC: Exploration and Discussion of Recent Advances}.
\newblock \bibinfo{journal}{\emph{arXiv preprint arXiv:2309.14329}} (\bibinfo{year}{2023}).
\newblock


\bibitem[He et~al\mbox{.}(2022)]%
        {he2022latent}
\bibfield{author}{\bibinfo{person}{Yingqing He}, \bibinfo{person}{Tianyu Yang}, \bibinfo{person}{Yong Zhang}, \bibinfo{person}{Ying Shan}, {and} \bibinfo{person}{Qifeng Chen}.} \bibinfo{year}{2022}\natexlab{}.
\newblock \showarticletitle{Latent video diffusion models for high-fidelity long video generation}.
\newblock \bibinfo{journal}{\emph{arXiv preprint arXiv:2211.13221}} (\bibinfo{year}{2022}).
\newblock


\bibitem[Hessel et~al\mbox{.}(2021)]%
        {CLIPScore}
\bibfield{author}{\bibinfo{person}{Jack Hessel}, \bibinfo{person}{Ari Holtzman}, \bibinfo{person}{Maxwell Forbes}, \bibinfo{person}{Ronan Le~Bras}, {and} \bibinfo{person}{Yejin Choi}.} \bibinfo{year}{2021}\natexlab{}.
\newblock \showarticletitle{CLIPScore: A Reference-free Evaluation Metric for Image Captioning}. In \bibinfo{booktitle}{\emph{Proceedings of the 2021 Conference on Empirical Methods in Natural Language Processing (EMNLP)}}. \bibinfo{pages}{7514--7528}.
\newblock


\bibitem[Heusel et~al\mbox{.}(2017)]%
        {FID}
\bibfield{author}{\bibinfo{person}{Martin Heusel}, \bibinfo{person}{Hubert Ramsauer}, \bibinfo{person}{Thomas Unterthiner}, \bibinfo{person}{Bernhard Nessler}, {and} \bibinfo{person}{Sepp Hochreiter}.} \bibinfo{year}{2017}\natexlab{}.
\newblock \showarticletitle{GANs Trained by a Two Time-Scale Update Rule Converge to a Local Nash Equilibrium}. In \bibinfo{booktitle}{\emph{Advances in Neural Information Processing Systems}}, \bibfield{editor}{\bibinfo{person}{I.~Guyon}, \bibinfo{person}{U.~Von Luxburg}, \bibinfo{person}{S.~Bengio}, \bibinfo{person}{H.~Wallach}, \bibinfo{person}{R.~Fergus}, \bibinfo{person}{S.~Vishwanathan}, {and} \bibinfo{person}{R.~Garnett}} (Eds.), Vol.~\bibinfo{volume}{30}. \bibinfo{publisher}{Curran Associates, Inc.}
\newblock


\bibitem[Ho et~al\mbox{.}(2020)]%
        {ho2020denoising}
\bibfield{author}{\bibinfo{person}{Jonathan Ho}, \bibinfo{person}{Ajay Jain}, {and} \bibinfo{person}{Pieter Abbeel}.} \bibinfo{year}{2020}\natexlab{}.
\newblock \showarticletitle{Denoising diffusion probabilistic models}.
\newblock \bibinfo{journal}{\emph{Advances in neural information processing systems}}  \bibinfo{volume}{33} (\bibinfo{year}{2020}), \bibinfo{pages}{6840--6851}.
\newblock


\bibitem[Ho et~al\mbox{.}(2022)]%
        {ho2022video}
\bibfield{author}{\bibinfo{person}{Jonathan Ho}, \bibinfo{person}{Tim Salimans}, \bibinfo{person}{Alexey Gritsenko}, \bibinfo{person}{William Chan}, \bibinfo{person}{Mohammad Norouzi}, {and} \bibinfo{person}{David~J Fleet}.} \bibinfo{year}{2022}\natexlab{}.
\newblock \showarticletitle{Video diffusion models}.
\newblock \bibinfo{journal}{\emph{Advances in Neural Information Processing Systems}}  \bibinfo{volume}{35} (\bibinfo{year}{2022}), \bibinfo{pages}{8633--8646}.
\newblock


\bibitem[Hong et~al\mbox{.}(2022)]%
        {hong2022cogvideo}
\bibfield{author}{\bibinfo{person}{Wenyi Hong}, \bibinfo{person}{Ming Ding}, \bibinfo{person}{Wendi Zheng}, \bibinfo{person}{Xinghan Liu}, {and} \bibinfo{person}{Jie Tang}.} \bibinfo{year}{2022}\natexlab{}.
\newblock \showarticletitle{CogVideo: Large-scale Pretraining for Text-to-Video Generation via Transformers}. In \bibinfo{booktitle}{\emph{The Eleventh International Conference on Learning Representations}}.
\newblock


\bibitem[Hosu et~al\mbox{.}(2017)]%
        {KoNViD_1k}
\bibfield{author}{\bibinfo{person}{Vlad Hosu}, \bibinfo{person}{Franz Hahn}, \bibinfo{person}{Mohsen Jenadeleh}, \bibinfo{person}{Hanhe Lin}, \bibinfo{person}{Hui Men}, \bibinfo{person}{Tam{\'a}s Szir{\'a}nyi}, \bibinfo{person}{Shujun Li}, {and} \bibinfo{person}{Dietmar Saupe}.} \bibinfo{year}{2017}\natexlab{}.
\newblock \showarticletitle{The Konstanz natural video database (KoNViD-1k)}. In \bibinfo{booktitle}{\emph{2017 Ninth international conference on quality of multimedia experience (QoMEX)}}. IEEE, \bibinfo{pages}{1--6}.
\newblock


\bibitem[Hosu et~al\mbox{.}(2020)]%
        {KonIQ_10k}
\bibfield{author}{\bibinfo{person}{Vlad Hosu}, \bibinfo{person}{Hanhe Lin}, \bibinfo{person}{Tamas Sziranyi}, {and} \bibinfo{person}{Dietmar Saupe}.} \bibinfo{year}{2020}\natexlab{}.
\newblock \showarticletitle{KonIQ-10k: An ecologically valid database for deep learning of blind image quality assessment}.
\newblock \bibinfo{journal}{\emph{IEEE Transactions on Image Processing}}  \bibinfo{volume}{29} (\bibinfo{year}{2020}), \bibinfo{pages}{4041--4056}.
\newblock


\bibitem[Huang et~al\mbox{.}(2024)]%
        {VBench}
\bibfield{author}{\bibinfo{person}{Ziqi Huang}, \bibinfo{person}{Yinan He}, \bibinfo{person}{Jiashuo Yu}, \bibinfo{person}{Fan Zhang}, \bibinfo{person}{Chenyang Si}, \bibinfo{person}{Yuming Jiang}, \bibinfo{person}{Yuanhan Zhang}, \bibinfo{person}{Tianxing Wu}, \bibinfo{person}{Qingyang Jin}, \bibinfo{person}{Nattapol Chanpaisit}, \bibinfo{person}{Yaohui Wang}, \bibinfo{person}{Xinyuan Chen}, \bibinfo{person}{Limin Wang}, \bibinfo{person}{Dahua Lin}, \bibinfo{person}{Yu Qiao}, {and} \bibinfo{person}{Ziwei Liu}.} \bibinfo{year}{2024}\natexlab{}.
\newblock \showarticletitle{VBench: Comprehensive Benchmark Suite for Video Generative Models}. In \bibinfo{booktitle}{\emph{Proceedings of the IEEE/CVF Conference on Computer Vision and Pattern Recognition (CVPR)}}. \bibinfo{pages}{21807--21818}.
\newblock


\bibitem[Hudson and Manning(2019)]%
        {GQA}
\bibfield{author}{\bibinfo{person}{David~A. Hudson} {and} \bibinfo{person}{Christopher~D. Manning}.} \bibinfo{year}{2019}\natexlab{}.
\newblock \showarticletitle{GQA: Visual Question Answering with Graph-Structured Scenes}. In \bibinfo{booktitle}{\emph{Proceedings of the IEEE/CVF Conference on Computer Vision and Pattern Recognition (CVPR)}}. \bibinfo{pages}{6706--6715}.
\newblock


\bibitem[Ioffe and Szegedy(2015)]%
        {inception}
\bibfield{author}{\bibinfo{person}{Sergey Ioffe} {and} \bibinfo{person}{Christian Szegedy}.} \bibinfo{year}{2015}\natexlab{}.
\newblock \showarticletitle{Batch Normalization: Accelerating Deep Network Training by Reducing Internal Covariate Shift}. In \bibinfo{booktitle}{\emph{Proceedings of the 32nd International Conference on Machine Learning}} \emph{(\bibinfo{series}{Proceedings of Machine Learning Research}, Vol.~\bibinfo{volume}{37})}, \bibfield{editor}{\bibinfo{person}{Francis Bach} {and} \bibinfo{person}{David Blei}} (Eds.). \bibinfo{publisher}{PMLR}, \bibinfo{address}{Lille, France}, \bibinfo{pages}{448--456}.
\newblock


\bibitem[Ke et~al\mbox{.}(2021)]%
        {MUSIQ}
\bibfield{author}{\bibinfo{person}{Junjie Ke}, \bibinfo{person}{Qifei Wang}, \bibinfo{person}{Yilin Wang}, \bibinfo{person}{Peyman Milanfar}, {and} \bibinfo{person}{Feng Yang}.} \bibinfo{year}{2021}\natexlab{}.
\newblock \showarticletitle{MUSIQ: Multi-scale Image Quality Transformer}. In \bibinfo{booktitle}{\emph{2021 IEEE/CVF International Conference on Computer Vision (ICCV)}}. \bibinfo{pages}{5128--5137}.
\newblock
\urldef\tempurl%
\url{https://doi.org/10.1109/ICCV48922.2021.00510}
\showDOI{\tempurl}


\bibitem[Khachatryan et~al\mbox{.}(2023)]%
        {Text2Video}
\bibfield{author}{\bibinfo{person}{Levon Khachatryan}, \bibinfo{person}{Andranik Movsisyan}, \bibinfo{person}{Vahram Tadevosyan}, \bibinfo{person}{Roberto Henschel}, \bibinfo{person}{Zhangyang Wang}, \bibinfo{person}{Shant Navasardyan}, {and} \bibinfo{person}{Humphrey Shi}.} \bibinfo{year}{2023}\natexlab{}.
\newblock \bibinfo{title}{Text2Video-Zero: Text-to-Image Diffusion Models are Zero-Shot Video Generators}.
\newblock
\newblock
\showeprint[arxiv]{2303.13439}~[cs.CV]


\bibitem[Kirstain et~al\mbox{.}(2023)]%
        {PickScore}
\bibfield{author}{\bibinfo{person}{Yuval Kirstain}, \bibinfo{person}{Adam Poliak}, \bibinfo{person}{Uriel Singer}, {and} \bibinfo{person}{Omer Levy}.} \bibinfo{year}{2023}\natexlab{}.
\newblock \showarticletitle{Pick-a-Pic: An Open Dataset of User Preferences for Text-to-Image Generation}. In \bibinfo{booktitle}{\emph{Advances in Neural Information Processing Systems}}, Vol.~\bibinfo{volume}{36}.
\newblock


\bibitem[Korhonen(2019)]%
        {TLVQM}
\bibfield{author}{\bibinfo{person}{Jari Korhonen}.} \bibinfo{year}{2019}\natexlab{}.
\newblock \showarticletitle{Two-Level Approach for No-Reference Consumer Video Quality Assessment}.
\newblock \bibinfo{journal}{\emph{IEEE Transactions on Image Processing}} \bibinfo{volume}{28}, \bibinfo{number}{12} (\bibinfo{year}{2019}), \bibinfo{pages}{5923--5938}.
\newblock
\urldef\tempurl%
\url{https://doi.org/10.1109/TIP.2019.2923051}
\showDOI{\tempurl}


\bibitem[Kou et~al\mbox{.}(2024)]%
        {T2VQA}
\bibfield{author}{\bibinfo{person}{Tengchuan Kou}, \bibinfo{person}{Xiaohong Liu}, \bibinfo{person}{Zicheng Zhang}, \bibinfo{person}{Chunyi Li}, \bibinfo{person}{Haoning Wu}, \bibinfo{person}{Xiongkuo Min}, \bibinfo{person}{Guangtao Zhai}, {and} \bibinfo{person}{Ning Liu}.} \bibinfo{year}{2024}\natexlab{}.
\newblock \showarticletitle{Subjective-aligned dataset and metric for text-to-video quality assessment}. In \bibinfo{booktitle}{\emph{Proceedings of the 32nd ACM International Conference on Multimedia}}. \bibinfo{pages}{7793--7802}.
\newblock


\bibitem[Li et~al\mbox{.}(2024)]%
        {AIGIQA_20K}
\bibfield{author}{\bibinfo{person}{Chunyi Li}, \bibinfo{person}{Tengchuan Kou}, \bibinfo{person}{Yixuan Gao}, \bibinfo{person}{Yuqin Cao}, \bibinfo{person}{Wei Sun}, \bibinfo{person}{Zicheng Zhang}, \bibinfo{person}{Yingjie Zhou}, \bibinfo{person}{Zhichao Zhang}, \bibinfo{person}{Weixia Zhang}, \bibinfo{person}{Haoning Wu}, \bibinfo{person}{Xiaohong Liu}, \bibinfo{person}{Xiongkuo Min}, {and} \bibinfo{person}{Guangtao Zhai}.} \bibinfo{year}{2024}\natexlab{}.
\newblock \showarticletitle{AIGIQA-20K: A Large Database for AI-Generated Image Quality Assessment}. In \bibinfo{booktitle}{\emph{2024 IEEE/CVF Conference on Computer Vision and Pattern Recognition Workshops (CVPRW)}}. \bibinfo{pages}{6327--6336}.
\newblock
\urldef\tempurl%
\url{https://doi.org/10.1109/CVPRW63382.2024.00636}
\showDOI{\tempurl}


\bibitem[Li et~al\mbox{.}(2023a)]%
        {AGIQA_3K}
\bibfield{author}{\bibinfo{person}{Chunyi Li}, \bibinfo{person}{Zicheng Zhang}, \bibinfo{person}{Haoning Wu}, \bibinfo{person}{Wei Sun}, \bibinfo{person}{Xiongkuo Min}, \bibinfo{person}{Xiaohong Liu}, \bibinfo{person}{Guangtao Zhai}, {and} \bibinfo{person}{Weisi Lin}.} \bibinfo{year}{2023}\natexlab{a}.
\newblock \showarticletitle{AGIQA-3K: An Open Database for AI-Generated Image Quality Assessment}.
\newblock \bibinfo{journal}{\emph{IEEE Transactions on Circuits and Systems for Video Technology}} (\bibinfo{year}{2023}), \bibinfo{pages}{1--1}.
\newblock
\urldef\tempurl%
\url{https://doi.org/10.1109/TCSVT.2023.3319020}
\showDOI{\tempurl}


\bibitem[Li et~al\mbox{.}(2019)]%
        {VSFA}
\bibfield{author}{\bibinfo{person}{Dingquan Li}, \bibinfo{person}{Tingting Jiang}, {and} \bibinfo{person}{Ming Jiang}.} \bibinfo{year}{2019}\natexlab{}.
\newblock \showarticletitle{Quality Assessment of In-the-Wild Videos}. In \bibinfo{booktitle}{\emph{Proceedings of the 27th ACM International Conference on Multimedia}} (Nice, France) \emph{(\bibinfo{series}{MM '19})}. \bibinfo{publisher}{Association for Computing Machinery}, \bibinfo{address}{New York, NY, USA}, \bibinfo{pages}{2351–2359}.
\newblock
\showISBNx{9781450368896}
\urldef\tempurl%
\url{https://doi.org/10.1145/3343031.3351028}
\showDOI{\tempurl}


\bibitem[Li et~al\mbox{.}(2022)]%
        {BLIP}
\bibfield{author}{\bibinfo{person}{Junnan Li}, \bibinfo{person}{Dongxu Li}, \bibinfo{person}{Caiming Xiong}, {and} \bibinfo{person}{Steven Hoi}.} \bibinfo{year}{2022}\natexlab{}.
\newblock \showarticletitle{BLIP: Bootstrapping Language-Image Pre-training for Unified Vision-Language Understanding and Generation}. In \bibinfo{booktitle}{\emph{ICML}}.
\newblock


\bibitem[Li et~al\mbox{.}(2018)]%
        {li2018video}
\bibfield{author}{\bibinfo{person}{Yitong Li}, \bibinfo{person}{Martin Min}, \bibinfo{person}{Dinghan Shen}, \bibinfo{person}{David Carlson}, {and} \bibinfo{person}{Lawrence Carin}.} \bibinfo{year}{2018}\natexlab{}.
\newblock \showarticletitle{Video generation from text}. In \bibinfo{booktitle}{\emph{Proceedings of the AAAI conference on artificial intelligence}}, Vol.~\bibinfo{volume}{32}.
\newblock


\bibitem[Li et~al\mbox{.}(2023b)]%
        {Amt}
\bibfield{author}{\bibinfo{person}{Zhen Li}, \bibinfo{person}{Zuo-Liang Zhu}, \bibinfo{person}{Ling-Hao Han}, \bibinfo{person}{Qibin Hou}, \bibinfo{person}{Chun-Le Guo}, {and} \bibinfo{person}{Ming-Ming Cheng}.} \bibinfo{year}{2023}\natexlab{b}.
\newblock \showarticletitle{Amt: All-pairs multi-field transforms for efficient frame interpolation}. In \bibinfo{booktitle}{\emph{Proceedings of the IEEE/CVF Conference on Computer Vision and Pattern Recognition}}. \bibinfo{pages}{9801--9810}.
\newblock


\bibitem[Liang et~al\mbox{.}(2022)]%
        {liang2022nuwa}
\bibfield{author}{\bibinfo{person}{Jian Liang}, \bibinfo{person}{Chenfei Wu}, \bibinfo{person}{Xiaowei Hu}, \bibinfo{person}{Zhe Gan}, \bibinfo{person}{Jianfeng Wang}, \bibinfo{person}{Lijuan Wang}, \bibinfo{person}{Zicheng Liu}, \bibinfo{person}{Yuejian Fang}, {and} \bibinfo{person}{Nan Duan}.} \bibinfo{year}{2022}\natexlab{}.
\newblock \showarticletitle{Nuwa-infinity: Autoregressive over autoregressive generation for infinite visual synthesis}.
\newblock \bibinfo{journal}{\emph{Advances in Neural Information Processing Systems}}  \bibinfo{volume}{35} (\bibinfo{year}{2022}), \bibinfo{pages}{15420--15432}.
\newblock


\bibitem[Lin et~al\mbox{.}(2014)]%
        {MSCOCO}
\bibfield{author}{\bibinfo{person}{Tsung-Yi Lin}, \bibinfo{person}{Zhihong Ma}, \bibinfo{person}{Peter~N Belhumeur}, {et~al\mbox{.}}} \bibinfo{year}{2014}\natexlab{}.
\newblock \showarticletitle{Microsoft coco: Common objects in context}. In \bibinfo{booktitle}{\emph{European conference on computer vision}}. Springer, \bibinfo{pages}{740--755}.
\newblock


\bibitem[Lin et~al\mbox{.}(2025)]%
        {VQAScore}
\bibfield{author}{\bibinfo{person}{Zhiqiu Lin}, \bibinfo{person}{Deepak Pathak}, \bibinfo{person}{Baiqi Li}, \bibinfo{person}{Jiayao Li}, \bibinfo{person}{Xide Xia}, \bibinfo{person}{Graham Neubig}, \bibinfo{person}{Pengchuan Zhang}, {and} \bibinfo{person}{Deva Ramanan}.} \bibinfo{year}{2025}\natexlab{}.
\newblock \showarticletitle{Evaluating Text-to-Visual Generation with Image-to-Text Generation}. In \bibinfo{booktitle}{\emph{Computer Vision -- ECCV 2024}}, \bibfield{editor}{\bibinfo{person}{Ale{\v{s}} Leonardis}, \bibinfo{person}{Elisa Ricci}, \bibinfo{person}{Stefan Roth}, \bibinfo{person}{Olga Russakovsky}, \bibinfo{person}{Torsten Sattler}, {and} \bibinfo{person}{G{\"u}l Varol}} (Eds.). \bibinfo{publisher}{Springer Nature Switzerland}, \bibinfo{address}{Cham}, \bibinfo{pages}{366--384}.
\newblock
\showISBNx{978-3-031-72673-6}


\bibitem[Liu et~al\mbox{.}(2024)]%
        {EvalCrafter}
\bibfield{author}{\bibinfo{person}{Yaofang Liu}, \bibinfo{person}{Xiaodong Cun}, \bibinfo{person}{Xuebo Liu}, \bibinfo{person}{Xintao Wang}, \bibinfo{person}{Yong Zhang}, \bibinfo{person}{Haoxin Chen}, \bibinfo{person}{Yang Liu}, \bibinfo{person}{Tieyong Zeng}, \bibinfo{person}{Raymond Chan}, {and} \bibinfo{person}{Ying Shan}.} \bibinfo{year}{2024}\natexlab{}.
\newblock \showarticletitle{EvalCrafter: Benchmarking and Evaluating Large Video Generation Models}. In \bibinfo{booktitle}{\emph{Proceedings of the IEEE/CVF Conference on Computer Vision and Pattern Recognition (CVPR)}}. \bibinfo{pages}{22139--22149}.
\newblock


\bibitem[Liu et~al\mbox{.}(2023)]%
        {FETV}
\bibfield{author}{\bibinfo{person}{Yuanxin Liu}, \bibinfo{person}{Lei Li}, \bibinfo{person}{Shuhuai Ren}, \bibinfo{person}{Rundong Gao}, \bibinfo{person}{Shicheng Li}, \bibinfo{person}{Sishuo Chen}, \bibinfo{person}{Xu Sun}, {and} \bibinfo{person}{Lu Hou}.} \bibinfo{year}{2023}\natexlab{}.
\newblock \showarticletitle{FETV: A Benchmark for Fine-Grained Evaluation of Open-Domain Text-to-Video Generation}. In \bibinfo{booktitle}{\emph{Advances in Neural Information Processing Systems}}, \bibfield{editor}{\bibinfo{person}{A.~Oh}, \bibinfo{person}{T.~Naumann}, \bibinfo{person}{A.~Globerson}, \bibinfo{person}{K.~Saenko}, \bibinfo{person}{M.~Hardt}, {and} \bibinfo{person}{S.~Levine}} (Eds.), Vol.~\bibinfo{volume}{36}. \bibinfo{publisher}{Curran Associates, Inc.}, \bibinfo{pages}{62352--62387}.
\newblock


\bibitem[Liu et~al\mbox{.}(2021)]%
        {Swin}
\bibfield{author}{\bibinfo{person}{Ze Liu}, \bibinfo{person}{Yutong Lin}, \bibinfo{person}{Yue Cao}, \bibinfo{person}{Han Hu}, \bibinfo{person}{Yixuan Wei}, \bibinfo{person}{Zheng Zhang}, \bibinfo{person}{Stephen Lin}, {and} \bibinfo{person}{Baining Guo}.} \bibinfo{year}{2021}\natexlab{}.
\newblock \showarticletitle{Swin Transformer: Hierarchical Vision Transformer using Shifted Windows}. In \bibinfo{booktitle}{\emph{Proceedings of the IEEE/CVF International Conference on Computer Vision (ICCV)}}.
\newblock


\bibitem[Luo et~al\mbox{.}(2023)]%
        {VideoFusion}
\bibfield{author}{\bibinfo{person}{Zhengxiong Luo}, \bibinfo{person}{Dayou Chen}, \bibinfo{person}{Yingya Zhang}, \bibinfo{person}{Yan Huang}, \bibinfo{person}{Liang Wang}, \bibinfo{person}{Yujun Shen}, \bibinfo{person}{Deli Zhao}, \bibinfo{person}{Jingren Zhou}, {and} \bibinfo{person}{Tieniu Tan}.} \bibinfo{year}{2023}\natexlab{}.
\newblock \showarticletitle{Notice of Removal: VideoFusion: Decomposed Diffusion Models for High-Quality Video Generation}. In \bibinfo{booktitle}{\emph{2023 IEEE/CVF Conference on Computer Vision and Pattern Recognition (CVPR)}}. \bibinfo{pages}{10209--10218}.
\newblock
\urldef\tempurl%
\url{https://doi.org/10.1109/CVPR52729.2023.00984}
\showDOI{\tempurl}


\bibitem[Min et~al\mbox{.}(2024)]%
        {min2024perceptual}
\bibfield{author}{\bibinfo{person}{Xiongkuo Min}, \bibinfo{person}{Huiyu Duan}, \bibinfo{person}{Wei Sun}, \bibinfo{person}{Yucheng Zhu}, {and} \bibinfo{person}{Guangtao Zhai}.} \bibinfo{year}{2024}\natexlab{}.
\newblock \showarticletitle{Perceptual Video Quality Assessment: A Survey}.
\newblock \bibinfo{journal}{\emph{arXiv preprint arXiv:2402.03413}} (\bibinfo{year}{2024}).
\newblock


\bibitem[Mittal et~al\mbox{.}(2012a)]%
        {BRISQUE}
\bibfield{author}{\bibinfo{person}{Anish Mittal}, \bibinfo{person}{Anush~Krishna Moorthy}, {and} \bibinfo{person}{Alan~Conrad Bovik}.} \bibinfo{year}{2012}\natexlab{a}.
\newblock \showarticletitle{No-reference image quality assessment in the spatial domain}.
\newblock \bibinfo{journal}{\emph{IEEE Transactions on image processing}} \bibinfo{volume}{21}, \bibinfo{number}{12} (\bibinfo{year}{2012}), \bibinfo{pages}{4695--4708}.
\newblock


\bibitem[Mittal et~al\mbox{.}(2012b)]%
        {NIQE}
\bibfield{author}{\bibinfo{person}{Anish Mittal}, \bibinfo{person}{Rajiv Soundararajan}, {and} \bibinfo{person}{Alan~C Bovik}.} \bibinfo{year}{2012}\natexlab{b}.
\newblock \showarticletitle{Making a “completely blind” image quality analyzer}.
\newblock \bibinfo{journal}{\emph{IEEE Signal processing letters}} \bibinfo{volume}{20}, \bibinfo{number}{3} (\bibinfo{year}{2012}), \bibinfo{pages}{209--212}.
\newblock


\bibitem[Mullan et~al\mbox{.}(2023)]%
        {Hotshot}
\bibfield{author}{\bibinfo{person}{John Mullan}, \bibinfo{person}{Duncan Crawbuck}, {and} \bibinfo{person}{Aakash Sastry}.} \bibinfo{year}{2023}\natexlab{}.
\newblock \bibinfo{booktitle}{\emph{{Hotshot-XL}}}.
\newblock
\urldef\tempurl%
\url{https://github.com/hotshotco/hotshot-xl}
\showURL{%
\tempurl}


\bibitem[Otani et~al\mbox{.}(2023)]%
        {toward}
\bibfield{author}{\bibinfo{person}{Mayu Otani}, \bibinfo{person}{Riku Togashi}, \bibinfo{person}{Yu Sawai}, \bibinfo{person}{Ryosuke Ishigami}, \bibinfo{person}{Yuta Nakashima}, \bibinfo{person}{Esa Rahtu}, \bibinfo{person}{Janne Heikkilä}, {and} \bibinfo{person}{Shin'ichi Satoh}.} \bibinfo{year}{2023}\natexlab{}.
\newblock \showarticletitle{Toward Verifiable and Reproducible Human Evaluation for Text-to-Image Generation}.
\newblock  (\bibinfo{year}{2023}), \bibinfo{pages}{14277--14286}.
\newblock
\urldef\tempurl%
\url{https://doi.org/10.1109/CVPR52729.2023.01372}
\showDOI{\tempurl}


\bibitem[Peng et~al\mbox{.}(2024)]%
        {IPCE}
\bibfield{author}{\bibinfo{person}{Fei Peng}, \bibinfo{person}{Huiyuan Fu}, \bibinfo{person}{Anlong Ming}, \bibinfo{person}{Chuanming Wang}, \bibinfo{person}{Huadong Ma}, \bibinfo{person}{Shuai He}, \bibinfo{person}{Zifei Dou}, {and} \bibinfo{person}{Shu Chen}.} \bibinfo{year}{2024}\natexlab{}.
\newblock \showarticletitle{AIGC Image Quality Assessment via Image-Prompt Correspondence}. In \bibinfo{booktitle}{\emph{2024 IEEE/CVF Conference on Computer Vision and Pattern Recognition Workshops (CVPRW)}}. \bibinfo{pages}{6432--6441}.
\newblock
\urldef\tempurl%
\url{https://doi.org/10.1109/CVPRW63382.2024.00644}
\showDOI{\tempurl}


\bibitem[Radford et~al\mbox{.}(2021)]%
        {CLIP}
\bibfield{author}{\bibinfo{person}{A. Radford}, \bibinfo{person}{J.~W. Kim}, \bibinfo{person}{C. Hallacy}, \bibinfo{person}{A. Ramesh}, \bibinfo{person}{G. Goh}, \bibinfo{person}{S. Agarwal}, \bibinfo{person}{G. Sastry}, \bibinfo{person}{A. Askell}, \bibinfo{person}{P. Mishkin}, \bibinfo{person}{J. Clark}, \bibinfo{person}{G. Krueger}, {and} \bibinfo{person}{I. Sutskever}.} \bibinfo{year}{2021}\natexlab{}.
\newblock \showarticletitle{Learning Transferable Visual Models From Natural Language Supervision}.
\newblock \bibinfo{journal}{\emph{Proceedings of the 38th International Conference on Machine Learning}} (\bibinfo{year}{2021}).
\newblock


\bibitem[Salimans et~al\mbox{.}(2016)]%
        {IS}
\bibfield{author}{\bibinfo{person}{Tim Salimans}, \bibinfo{person}{Ian Goodfellow}, \bibinfo{person}{Wojciech Zaremba}, \bibinfo{person}{Vicki Cheung}, \bibinfo{person}{Alec Radford}, {and} \bibinfo{person}{Xi Chen}.} \bibinfo{year}{2016}\natexlab{}.
\newblock \showarticletitle{Improved techniques for training gans}.
\newblock \bibinfo{journal}{\emph{Advances in neural information processing systems}}  \bibinfo{volume}{29} (\bibinfo{year}{2016}).
\newblock


\bibitem[Schuhmann et~al\mbox{.}(2021)]%
        {LAION_5B}
\bibfield{author}{\bibinfo{person}{Christoph Schuhmann} {et~al\mbox{.}}} \bibinfo{year}{2021}\natexlab{}.
\newblock \bibinfo{title}{LAION-5B: A New Dataset for CLIP-based Training and Beyond}.
\newblock \bibinfo{howpublished}{\url{https://laion.ai/}}.
\newblock
\newblock
\shownote{Accessed: 2024-12-18}.


\bibitem[Su et~al\mbox{.}(2020)]%
        {HyperIQA}
\bibfield{author}{\bibinfo{person}{Shaolin Su}, \bibinfo{person}{Qingsen Yan}, \bibinfo{person}{Yu Zhu}, \bibinfo{person}{Cheng Zhang}, \bibinfo{person}{Xin Ge}, \bibinfo{person}{Jinqiu Sun}, {and} \bibinfo{person}{Yanning Zhang}.} \bibinfo{year}{2020}\natexlab{}.
\newblock \showarticletitle{Blindly Assess Image Quality in the Wild Guided by a Self-Adaptive Hyper Network}. In \bibinfo{booktitle}{\emph{Proceedings of the IEEE/CVF Conference on Computer Vision and Pattern Recognition (CVPR)}}.
\newblock


\bibitem[Sun et~al\mbox{.}({[n.\,d.]})]%
        {sunsora}
\bibfield{author}{\bibinfo{person}{Rui Sun}, \bibinfo{person}{Yumin Zhang}, \bibinfo{person}{Tejal Shah}, \bibinfo{person}{Jiaohao Sun}, \bibinfo{person}{Shuoying Zhang}, \bibinfo{person}{Wenqi Li}, \bibinfo{person}{Haoran Duan}, {and} \bibinfo{person}{Bo Wei}.} \bibinfo{year}{[n.\,d.]}\natexlab{}.
\newblock \showarticletitle{From Sora What We Can See: A Survey of Text-to-Video Generation}.
\newblock  (\bibinfo{year}{[n.\,d.]}).
\newblock


\bibitem[Sun et~al\mbox{.}(2022)]%
        {simpleVQA}
\bibfield{author}{\bibinfo{person}{Wei Sun}, \bibinfo{person}{Xiongkuo Min}, \bibinfo{person}{Wei Lu}, {and} \bibinfo{person}{Guangtao Zhai}.} \bibinfo{year}{2022}\natexlab{}.
\newblock \showarticletitle{A Deep Learning Based No-Reference Quality Assessment Model for UGC Videos}. In \bibinfo{booktitle}{\emph{Proceedings of the 30th ACM International Conference on Multimedia}}. \bibinfo{pages}{856–865}.
\newblock


\bibitem[Sun et~al\mbox{.}(2023)]%
        {StairIQA}
\bibfield{author}{\bibinfo{person}{Wei Sun}, \bibinfo{person}{Xiongkuo Min}, \bibinfo{person}{Danyang Tu}, \bibinfo{person}{Siwei Ma}, {and} \bibinfo{person}{Guangtao Zhai}.} \bibinfo{year}{2023}\natexlab{}.
\newblock \showarticletitle{Blind quality assessment for in-the-wild images via hierarchical feature fusion and iterative mixed database training}.
\newblock \bibinfo{journal}{\emph{IEEE Journal of Selected Topics in Signal Processing}} (\bibinfo{year}{2023}).
\newblock


\bibitem[Sun et~al\mbox{.}(2024)]%
        {sun2024analysis}
\bibfield{author}{\bibinfo{person}{Wei Sun}, \bibinfo{person}{Wen Wen}, \bibinfo{person}{Xiongkuo Min}, \bibinfo{person}{Long Lan}, \bibinfo{person}{Guangtao Zhai}, {and} \bibinfo{person}{Kede Ma}.} \bibinfo{year}{2024}\natexlab{}.
\newblock \showarticletitle{Analysis of video quality datasets via design of minimalistic video quality models}.
\newblock \bibinfo{journal}{\emph{IEEE Transactions on Pattern Analysis and Machine Intelligence}} (\bibinfo{year}{2024}).
\newblock


\bibitem[Teed and Deng(2020)]%
        {RAFT}
\bibfield{author}{\bibinfo{person}{Zachary Teed} {and} \bibinfo{person}{Jia Deng}.} \bibinfo{year}{2020}\natexlab{}.
\newblock \showarticletitle{Raft: Recurrent all-pairs field transforms for optical flow}. In \bibinfo{booktitle}{\emph{Computer Vision--ECCV 2020: 16th European Conference, Glasgow, UK, August 23--28, 2020, Proceedings, Part II 16}}. Springer, \bibinfo{pages}{402--419}.
\newblock


\bibitem[Tu et~al\mbox{.}(2021a)]%
        {VIDEAL}
\bibfield{author}{\bibinfo{person}{Zhengzhong Tu}, \bibinfo{person}{Yilin Wang}, \bibinfo{person}{Neil Birkbeck}, \bibinfo{person}{Balu Adsumilli}, {and} \bibinfo{person}{Alan~C Bovik}.} \bibinfo{year}{2021}\natexlab{a}.
\newblock \showarticletitle{UGC-VQA: Benchmarking blind video quality assessment for user generated content}.
\newblock \bibinfo{journal}{\emph{IEEE Transactions on Image Processing}}  \bibinfo{volume}{30} (\bibinfo{year}{2021}), \bibinfo{pages}{4449--4464}.
\newblock


\bibitem[Tu et~al\mbox{.}(2021b)]%
        {RAPIQUE}
\bibfield{author}{\bibinfo{person}{Zhengzhong Tu}, \bibinfo{person}{Xiangxu Yu}, \bibinfo{person}{Yilin Wang}, \bibinfo{person}{Neil Birkbeck}, \bibinfo{person}{Balu Adsumilli}, {and} \bibinfo{person}{Alan~C. Bovik}.} \bibinfo{year}{2021}\natexlab{b}.
\newblock \showarticletitle{RAPIQUE: Rapid and Accurate Video Quality Prediction of User Generated Content}.
\newblock \bibinfo{journal}{\emph{IEEE Open Journal of Signal Processing}}  \bibinfo{volume}{2} (\bibinfo{year}{2021}), \bibinfo{pages}{425--440}.
\newblock
\urldef\tempurl%
\url{https://doi.org/10.1109/OJSP.2021.3090333}
\showDOI{\tempurl}


\bibitem[Unterthiner et~al\mbox{.}(2019)]%
        {FVD}
\bibfield{author}{\bibinfo{person}{Thomas Unterthiner}, \bibinfo{person}{Sjoerd van Steenkiste}, \bibinfo{person}{Karol Kurach}, \bibinfo{person}{Raphael Marinier}, \bibinfo{person}{Marcin Michalski}, {and} \bibinfo{person}{Sylvain Gelly}.} \bibinfo{year}{2019}\natexlab{}.
\newblock \bibinfo{title}{Towards Accurate Generative Models of Video: A New Metric Challenges}.
\newblock
\newblock
\showeprint[arxiv]{1812.01717}~[cs.CV]


\bibitem[Villegas et~al\mbox{.}(2022)]%
        {villegas2022phenaki}
\bibfield{author}{\bibinfo{person}{Ruben Villegas}, \bibinfo{person}{Mohammad Babaeizadeh}, \bibinfo{person}{Pieter-Jan Kindermans}, \bibinfo{person}{Hernan Moraldo}, \bibinfo{person}{Han Zhang}, \bibinfo{person}{Mohammad~Taghi Saffar}, \bibinfo{person}{Santiago Castro}, \bibinfo{person}{Julius Kunze}, {and} \bibinfo{person}{Dumitru Erhan}.} \bibinfo{year}{2022}\natexlab{}.
\newblock \showarticletitle{Phenaki: Variable length video generation from open domain textual descriptions}. In \bibinfo{booktitle}{\emph{International Conference on Learning Representations}}.
\newblock


\bibitem[Wang et~al\mbox{.}(2023a)]%
        {CLIP_IQA}
\bibfield{author}{\bibinfo{person}{Jianyi Wang}, \bibinfo{person}{Kelvin~C.K. Chan}, {and} \bibinfo{person}{Chen~Change Loy}.} \bibinfo{year}{2023}\natexlab{a}.
\newblock \showarticletitle{Exploring CLIP for Assessing the Look and Feel of Images}.
\newblock \bibinfo{journal}{\emph{Proceedings of the AAAI Conference on Artificial Intelligence}} \bibinfo{volume}{37}, \bibinfo{number}{2} (\bibinfo{date}{Jun.} \bibinfo{year}{2023}), \bibinfo{pages}{2555--2563}.
\newblock
\urldef\tempurl%
\url{https://doi.org/10.1609/aaai.v37i2.25353}
\showDOI{\tempurl}


\bibitem[Wang et~al\mbox{.}(2023c)]%
        {Videomae_v2}
\bibfield{author}{\bibinfo{person}{Limin Wang}, \bibinfo{person}{Bingkun Huang}, \bibinfo{person}{Zhiyu Zhao}, \bibinfo{person}{Zhan Tong}, \bibinfo{person}{Yinan He}, \bibinfo{person}{Yi Wang}, \bibinfo{person}{Yali Wang}, {and} \bibinfo{person}{Yu Qiao}.} \bibinfo{year}{2023}\natexlab{c}.
\newblock \showarticletitle{Videomae v2: Scaling video masked autoencoders with dual masking}. In \bibinfo{booktitle}{\emph{Proceedings of the IEEE/CVF Conference on Computer Vision and Pattern Recognition}}. \bibinfo{pages}{14549--14560}.
\newblock


\bibitem[Wang et~al\mbox{.}(2024)]%
        {MA_AGIQA}
\bibfield{author}{\bibinfo{person}{Puyi Wang}, \bibinfo{person}{Wei Sun}, \bibinfo{person}{Zicheng Zhang}, \bibinfo{person}{Jun Jia}, \bibinfo{person}{Yanwei Jiang}, \bibinfo{person}{Zhichao Zhang}, \bibinfo{person}{Xiongkuo Min}, {and} \bibinfo{person}{Guangtao Zhai}.} \bibinfo{year}{2024}\natexlab{}.
\newblock \showarticletitle{Large Multi-modality Model Assisted AI-Generated Image Quality Assessment}. In \bibinfo{booktitle}{\emph{Proceedings of the 32nd ACM International Conference on Multimedia}}. \bibinfo{pages}{7803--7812}.
\newblock


\bibitem[Wang et~al\mbox{.}(2023b)]%
        {Internvid}
\bibfield{author}{\bibinfo{person}{Yi Wang}, \bibinfo{person}{Yinan He}, \bibinfo{person}{Yizhuo Li}, \bibinfo{person}{Kunchang Li}, \bibinfo{person}{Jiashuo Yu}, \bibinfo{person}{Xin Ma}, \bibinfo{person}{Xinhao Li}, \bibinfo{person}{Guo Chen}, \bibinfo{person}{Xinyuan Chen}, \bibinfo{person}{Yaohui Wang}, {et~al\mbox{.}}} \bibinfo{year}{2023}\natexlab{b}.
\newblock \showarticletitle{Internvid: A large-scale video-text dataset for multimodal understanding and generation}.
\newblock \bibinfo{journal}{\emph{arXiv preprint arXiv:2307.06942}} (\bibinfo{year}{2023}).
\newblock


\bibitem[Wang et~al\mbox{.}(2022)]%
        {viCLIP}
\bibfield{author}{\bibinfo{person}{Yi Wang}, \bibinfo{person}{Kunchang Li}, \bibinfo{person}{Yizhuo Li}, \bibinfo{person}{Yinan He}, \bibinfo{person}{Bingkun Huang}, \bibinfo{person}{Zhiyu Zhao}, \bibinfo{person}{Hongjie Zhang}, \bibinfo{person}{Jilan Xu}, \bibinfo{person}{Yi Liu}, \bibinfo{person}{Zun Wang}, \bibinfo{person}{Sen Xing}, \bibinfo{person}{Guo Chen}, \bibinfo{person}{Junting Pan}, \bibinfo{person}{Jiashuo Yu}, \bibinfo{person}{Yali Wang}, \bibinfo{person}{Limin Wang}, {and} \bibinfo{person}{Yu Qiao}.} \bibinfo{year}{2022}\natexlab{}.
\newblock \showarticletitle{InternVideo: General Video Foundation Models via Generative and Discriminative Learning}.
\newblock \bibinfo{journal}{\emph{arXiv preprint arXiv:2212.03191}} (\bibinfo{year}{2022}).
\newblock


\bibitem[Wang et~al\mbox{.}(2023d)]%
        {aigc_game}
\bibfield{author}{\bibinfo{person}{Yuntao Wang}, \bibinfo{person}{Yanghe Pan}, \bibinfo{person}{Miao Yan}, \bibinfo{person}{Zhou Su}, {and} \bibinfo{person}{Tom~H Luan}.} \bibinfo{year}{2023}\natexlab{d}.
\newblock \showarticletitle{A survey on ChatGPT: AI-generated contents, challenges, and solutions}.
\newblock \bibinfo{journal}{\emph{IEEE Open Journal of the Computer Society}} (\bibinfo{year}{2023}).
\newblock


\bibitem[Wen and Wang(2021)]%
        {rank_loss}
\bibfield{author}{\bibinfo{person}{Shaoguo Wen} {and} \bibinfo{person}{Junle Wang}.} \bibinfo{year}{2021}\natexlab{}.
\newblock \showarticletitle{A strong baseline for image and video quality assessment}.
\newblock \bibinfo{journal}{\emph{arXiv preprint arXiv:2111.07104}} (\bibinfo{year}{2021}).
\newblock


\bibitem[Wu et~al\mbox{.}(2022b)]%
        {wu2022nuwa}
\bibfield{author}{\bibinfo{person}{Chenfei Wu}, \bibinfo{person}{Jian Liang}, \bibinfo{person}{Lei Ji}, \bibinfo{person}{Fan Yang}, \bibinfo{person}{Yuejian Fang}, \bibinfo{person}{Daxin Jiang}, {and} \bibinfo{person}{Nan Duan}.} \bibinfo{year}{2022}\natexlab{b}.
\newblock \showarticletitle{N{\"u}wa: Visual synthesis pre-training for neural visual world creation}. In \bibinfo{booktitle}{\emph{European conference on computer vision}}. Springer, \bibinfo{pages}{720--736}.
\newblock


\bibitem[Wu et~al\mbox{.}(2022a)]%
        {FAST_VQA}
\bibfield{author}{\bibinfo{person}{Haoning Wu}, \bibinfo{person}{Chaofeng Chen}, \bibinfo{person}{Jingwen Hou}, \bibinfo{person}{Liang Liao}, \bibinfo{person}{Annan Wang}, \bibinfo{person}{Wenxiu Sun}, \bibinfo{person}{Qiong Yan}, {and} \bibinfo{person}{Weisi Lin}.} \bibinfo{year}{2022}\natexlab{a}.
\newblock \showarticletitle{FAST-VQA: Efficient End-to-End Video Quality Assessment with Fragment Sampling}. In \bibinfo{booktitle}{\emph{Computer Vision – ECCV 2022: 17th European Conference, Tel Aviv, Israel, October 23–27, 2022, Proceedings, Part VI}} (Tel Aviv, Israel). \bibinfo{publisher}{Springer-Verlag}, \bibinfo{address}{Berlin, Heidelberg}, \bibinfo{pages}{538–554}.
\newblock
\showISBNx{978-3-031-20067-0}
\urldef\tempurl%
\url{https://doi.org/10.1007/978-3-031-20068-7_31}
\showDOI{\tempurl}


\bibitem[Wu et~al\mbox{.}(2023e)]%
        {dover}
\bibfield{author}{\bibinfo{person}{Haoning Wu}, \bibinfo{person}{Erli Zhang}, \bibinfo{person}{Liang Liao}, \bibinfo{person}{Chaofeng Chen}, \bibinfo{person}{Jingwen~Hou Hou}, \bibinfo{person}{Annan Wang}, \bibinfo{person}{Wenxiu~Sun Sun}, \bibinfo{person}{Qiong Yan}, {and} \bibinfo{person}{Weisi Lin}.} \bibinfo{year}{2023}\natexlab{e}.
\newblock \showarticletitle{Exploring Video Quality Assessment on User Generated Contents from Aesthetic and Technical Perspectives}. In \bibinfo{booktitle}{\emph{International Conference on Computer Vision (ICCV)}}.
\newblock


\bibitem[Wu et~al\mbox{.}(2023f)]%
        {qalign}
\bibfield{author}{\bibinfo{person}{Haoning Wu}, \bibinfo{person}{Zicheng Zhang}, \bibinfo{person}{Weixia Zhang}, \bibinfo{person}{Chaofeng Chen}, \bibinfo{person}{Liang Liao}, \bibinfo{person}{Chunyi Li}, \bibinfo{person}{Yixuan Gao}, \bibinfo{person}{Annan Wang}, \bibinfo{person}{Erli Zhang}, \bibinfo{person}{Wenxiu Sun}, {et~al\mbox{.}}} \bibinfo{year}{2023}\natexlab{f}.
\newblock \showarticletitle{Q-align: Teaching lmms for visual scoring via discrete text-defined levels}.
\newblock \bibinfo{journal}{\emph{arXiv preprint arXiv:2312.17090}} (\bibinfo{year}{2023}).
\newblock


\bibitem[Wu et~al\mbox{.}(2023a)]%
        {Tune-A-Video}
\bibfield{author}{\bibinfo{person}{Jay~Zhangjie Wu}, \bibinfo{person}{Yixiao Ge}, \bibinfo{person}{Xintao Wang}, \bibinfo{person}{Stan~Weixian Lei}, \bibinfo{person}{Yuchao Gu}, \bibinfo{person}{Yufei Shi}, \bibinfo{person}{Wynne Hsu}, \bibinfo{person}{Ying Shan}, \bibinfo{person}{Xiaohu Qie}, {and} \bibinfo{person}{Mike~Zheng Shou}.} \bibinfo{year}{2023}\natexlab{a}.
\newblock \showarticletitle{Tune-A-Video: One-Shot Tuning of Image Diffusion Models for Text-to-Video Generation}. In \bibinfo{booktitle}{\emph{Proceedings of the IEEE/CVF International Conference on Computer Vision (ICCV)}}. \bibinfo{pages}{7623--7633}.
\newblock


\bibitem[Wu et~al\mbox{.}(2023b)]%
        {wu2023tune}
\bibfield{author}{\bibinfo{person}{Jay~Zhangjie Wu}, \bibinfo{person}{Yixiao Ge}, \bibinfo{person}{Xintao Wang}, \bibinfo{person}{Stan~Weixian Lei}, \bibinfo{person}{Yuchao Gu}, \bibinfo{person}{Yufei Shi}, \bibinfo{person}{Wynne Hsu}, \bibinfo{person}{Ying Shan}, \bibinfo{person}{Xiaohu Qie}, {and} \bibinfo{person}{Mike~Zheng Shou}.} \bibinfo{year}{2023}\natexlab{b}.
\newblock \showarticletitle{Tune-a-video: One-shot tuning of image diffusion models for text-to-video generation}. In \bibinfo{booktitle}{\emph{Proceedings of the IEEE/CVF International Conference on Computer Vision}}. \bibinfo{pages}{7623--7633}.
\newblock


\bibitem[Wu et~al\mbox{.}(2023c)]%
        {HPSv2}
\bibfield{author}{\bibinfo{person}{Xiaoshi Wu}, \bibinfo{person}{Yiming Hao}, \bibinfo{person}{Keqiang Sun}, \bibinfo{person}{Yixiong Chen}, \bibinfo{person}{Feng Zhu}, \bibinfo{person}{Rui Zhao}, {and} \bibinfo{person}{Hongsheng Li}.} \bibinfo{year}{2023}\natexlab{c}.
\newblock \bibinfo{title}{Human Preference Score v2: A Solid Benchmark for Evaluating Human Preferences of Text-to-Image Synthesis}.
\newblock
\newblock
\showeprint[arxiv]{2306.09341}~[cs.CV]


\bibitem[Wu et~al\mbox{.}(2023d)]%
        {HPSv1}
\bibfield{author}{\bibinfo{person}{Xiaoshi Wu}, \bibinfo{person}{Keqiang Sun}, \bibinfo{person}{Feng Zhu}, \bibinfo{person}{Rui Zhao}, {and} \bibinfo{person}{Hongsheng Li}.} \bibinfo{year}{2023}\natexlab{d}.
\newblock \showarticletitle{Human Preference Score: Better Aligning Text-to-Image Models with Human Preference}. In \bibinfo{booktitle}{\emph{Proceedings of the IEEE/CVF International Conference on Computer Vision (ICCV)}}. \bibinfo{pages}{2096--2105}.
\newblock


\bibitem[Xu et~al\mbox{.}(2023)]%
        {ImageReward}
\bibfield{author}{\bibinfo{person}{Jiazheng Xu}, \bibinfo{person}{Xiao Liu}, \bibinfo{person}{Yuchen Wu}, \bibinfo{person}{Yuxuan Tong}, \bibinfo{person}{Qinkai Li}, \bibinfo{person}{Min Ding}, \bibinfo{person}{Jie Tang}, {and} \bibinfo{person}{Yuxiao Dong}.} \bibinfo{year}{2023}\natexlab{}.
\newblock \showarticletitle{ImageReward: Learning and Evaluating Human Preferences for Text-to-Image Generation}. In \bibinfo{booktitle}{\emph{Advances in Neural Information Processing Systems}}.
\newblock


\bibitem[Yin et~al\mbox{.}(2023)]%
        {yin2023nuwa}
\bibfield{author}{\bibinfo{person}{Shengming Yin}, \bibinfo{person}{Chenfei Wu}, \bibinfo{person}{Huan Yang}, \bibinfo{person}{Jianfeng Wang}, \bibinfo{person}{Xiaodong Wang}, \bibinfo{person}{Minheng Ni}, \bibinfo{person}{Zhengyuan Yang}, \bibinfo{person}{Linjie Li}, \bibinfo{person}{Shuguang Liu}, \bibinfo{person}{Fan Yang}, {et~al\mbox{.}}} \bibinfo{year}{2023}\natexlab{}.
\newblock \showarticletitle{Nuwa-xl: Diffusion over diffusion for extremely long video generation}.
\newblock \bibinfo{journal}{\emph{arXiv preprint arXiv:2303.12346}} (\bibinfo{year}{2023}).
\newblock


\bibitem[Ying et~al\mbox{.}(2021)]%
        {PatchVQ}
\bibfield{author}{\bibinfo{person}{Zhenqiang Ying}, \bibinfo{person}{Maniratnam Mandal}, \bibinfo{person}{Deepti Ghadiyaram}, {and} \bibinfo{person}{Alan Bovik}.} \bibinfo{year}{2021}\natexlab{}.
\newblock \showarticletitle{Patch-VQ: 'Patching Up' the Video Quality Problem}. In \bibinfo{booktitle}{\emph{Proceedings of the IEEE/CVF Conference on Computer Vision and Pattern Recognition (CVPR)}}. \bibinfo{pages}{14019--14029}.
\newblock


\bibitem[Young et~al\mbox{.}(2014)]%
        {Flickr30K}
\bibfield{author}{\bibinfo{person}{Peter Young}, \bibinfo{person}{Devendra Hazarika}, \bibinfo{person}{Soujanya Poria}, {and} \bibinfo{person}{Erik Cambria}.} \bibinfo{year}{2014}\natexlab{}.
\newblock \showarticletitle{Image captioning and visual question answering based on deep neural networks}. In \bibinfo{booktitle}{\emph{Proceedings of the IEEE conference on computer vision and pattern recognition}}. \bibinfo{pages}{2045--2054}.
\newblock


\bibitem[Zhang et~al\mbox{.}(2023c)]%
        {zhang2023text}
\bibfield{author}{\bibinfo{person}{Chenshuang Zhang}, \bibinfo{person}{Chaoning Zhang}, \bibinfo{person}{Mengchun Zhang}, {and} \bibinfo{person}{In~So Kweon}.} \bibinfo{year}{2023}\natexlab{c}.
\newblock \showarticletitle{Text-to-image diffusion model in generative ai: A survey}.
\newblock \bibinfo{journal}{\emph{arXiv preprint arXiv:2303.07909}} (\bibinfo{year}{2023}).
\newblock


\bibitem[Zhang and Li(2012)]%
        {SSIM}
\bibfield{author}{\bibinfo{person}{Lin Zhang} {and} \bibinfo{person}{Hongyu Li}.} \bibinfo{year}{2012}\natexlab{}.
\newblock \showarticletitle{SR-SIM: A fast and high performance IQA index based on spectral residual}. In \bibinfo{booktitle}{\emph{2012 19th IEEE international conference on image processing}}. IEEE, \bibinfo{pages}{1473--1476}.
\newblock


\bibitem[Zhang et~al\mbox{.}(2023a)]%
        {zhang2023adding}
\bibfield{author}{\bibinfo{person}{Lvmin Zhang}, \bibinfo{person}{Anyi Rao}, {and} \bibinfo{person}{Maneesh Agrawala}.} \bibinfo{year}{2023}\natexlab{a}.
\newblock \showarticletitle{Adding conditional control to text-to-image diffusion models}. In \bibinfo{booktitle}{\emph{Proceedings of the IEEE/CVF International Conference on Computer Vision}}. \bibinfo{pages}{3836--3847}.
\newblock


\bibitem[Zhang et~al\mbox{.}(2021)]%
        {UNIQUE}
\bibfield{author}{\bibinfo{person}{Weixia Zhang}, \bibinfo{person}{Kede Ma}, \bibinfo{person}{Guangtao Zhai}, {and} \bibinfo{person}{Xiaokang Yang}.} \bibinfo{year}{2021}\natexlab{}.
\newblock \showarticletitle{Uncertainty-Aware Blind Image Quality Assessment in the Laboratory and Wild}.
\newblock \bibinfo{journal}{\emph{IEEE Transactions on Image Processing}}  \bibinfo{volume}{30} (\bibinfo{year}{2021}), \bibinfo{pages}{3474--3486}.
\newblock
\urldef\tempurl%
\url{https://doi.org/10.1109/TIP.2021.3061932}
\showDOI{\tempurl}


\bibitem[Zhang et~al\mbox{.}(2023b)]%
        {LIQE}
\bibfield{author}{\bibinfo{person}{Weixia Zhang}, \bibinfo{person}{Guangtao Zhai}, \bibinfo{person}{Ying Wei}, \bibinfo{person}{Xiaokang Yang}, {and} \bibinfo{person}{Kede Ma}.} \bibinfo{year}{2023}\natexlab{b}.
\newblock \showarticletitle{Blind Image Quality Assessment via Vision-Language Correspondence: A Multitask Learning Perspective}. In \bibinfo{booktitle}{\emph{Proceedings of the IEEE/CVF Conference on Computer Vision and Pattern Recognition (CVPR)}}. \bibinfo{pages}{14071--14081}.
\newblock


\bibitem[Zhong et~al\mbox{.}(2021)]%
        {WIT}
\bibfield{author}{\bibinfo{person}{Zhiwei Zhong}, \bibinfo{person}{Wen-Ting Hsu}, \bibinfo{person}{He Xu}, \bibinfo{person}{Tsung-Yi Lee}, \bibinfo{person}{Yung-Hsiang Chou}, \bibinfo{person}{Jan-Yu Lee}, \bibinfo{person}{Yi Yu}, \bibinfo{person}{Zhe Yang}, \bibinfo{person}{Chen Sun}, {et~al\mbox{.}}} \bibinfo{year}{2021}\natexlab{}.
\newblock \showarticletitle{WIT: Web-Image Text Pretraining for Cross-Modal Vision-Language Understanding}.
\newblock \bibinfo{journal}{\emph{arXiv preprint arXiv:2102.05246}} (\bibinfo{year}{2021}).
\newblock


\end{thebibliography}

\end{document}